\documentclass{article}

 \usepackage[main, final]{neurips_2026}

\usepackage[utf8]{inputenc} %
\usepackage[T1]{fontenc}    %
\usepackage{hyperref}       %
\usepackage{url}            %
\usepackage{booktabs}       %
\usepackage{amsfonts}       %
\usepackage{nicefrac}       %
\usepackage{microtype}      %
\usepackage{xcolor}         %

\usepackage{graphicx}
\usepackage{booktabs}
\usepackage{microtype}
\usepackage{subcaption}
\usepackage{booktabs} 
\usepackage{xcolor}
\usepackage{makecell}
\usepackage{multirow}
\usepackage{wrapfig}
\usepackage{adjustbox}
\usepackage[table]{xcolor}
\usepackage{tabularx}
\usepackage{microtype}
\definecolor{posecolor}{HTML}{E5F5E0}
\definecolor{trajcolor}{HTML}{E3F2FD}
\definecolor{bothcolor}{HTML}{FFF8E1}
\definecolor{highlightgray}{gray}{0.9}
\usepackage{amsmath}
\usepackage{xspace}
\newcommand{\eg}{\emph{e.g.}\xspace}
\newcommand{\ie}{\emph{i.e.}\xspace}
\usepackage{enumitem}
\newcommand{\STAB}[1]{\begin{tabular}{@{}c@{}}#1\end{tabular}}

\title{SimpliHuMoN: Simplifying Human Motion Prediction}

\author{%
  Aadya Agrawal \\
  University of Illinois Urbana-Champaign\\
  \texttt{aadyaa3@illinois.edu} \\
  \And
  Alexander Schwing \\
  University of Illinois Urbana-Champaign\\
  \texttt{aschwing@illinois.edu} \\
}

\begin{document}

\maketitle

\begin{abstract}
  Human motion prediction combines the tasks of trajectory forecasting and human pose prediction. For each of the two tasks, specialized models have been developed. Combining these models for holistic human motion prediction is non-trivial, and recent methods often face tradeoffs on established benchmarks for individual tasks. To address this, we propose a simple yet effective transformer-based model for human motion prediction. The model employs a stack of self-attention modules to effectively capture both spatial dependencies within a pose and temporal relationships across a motion sequence. This simple, streamlined, end-to-end model is sufficiently versatile to handle pose-only, trajectory-only, and combined prediction tasks without task-specific modifications. We demonstrate that this approach achieves state-of-the-art or on-par results across all tasks through extensive experiments on a wide range of benchmark datasets, including Human3.6M, AMASS, ETH-UCY, and 3DPW. We open-source our code at: \url{https://github.com/aadya-agrawal/SimpliHuMoN}.
\end{abstract}

\section{Introduction}
\label{sec:intro}

Human motion prediction, the task of forecasting future 3D human motion from a sequence of past observations, is a critical challenge with wide-ranging applications in autonomous driving~\citep{zheng2022multimodal, paden2016survey}, robotics~\citep{ZOU2024127683, salzmann2023robots}, virtual reality~\citep{clark2020system, fu2020capture, ro2019display}, and sports analytics~\citep{li2021baseball}. Because human motion is inherently multi-dimensional, non-linear, and ambiguous, the literature has largely tackled prediction of human motion by addressing distinct tasks individually: trajectory prediction~\citep{gu2022stochastic, bae2022learning, shi2023trajectory, bae2024singulartrajectory, yao2024trajclip, fang2025neuralized} and pose prediction~\citep{dang2022diverse, barquero2023belfusion, sun2024comusion, hosseininejad2025motionmap, curreli2025nonisotropic, xu2024learning}.

While making individual tasks easier to address, this differentiation also creates a gap: tasks such as pose and trajectory prediction are fundamentally interrelated and governed by the same underlying dynamics~\citep{zheng2025efficientmultipersonmotionprediction}, yet they are modeled separately using task-specific architectures. This has led to the development of specialized models that excel at one task but struggle to generalize, limiting their applicability. Notable exceptions that jointly model these different tasks, particularly in the context of multi-person motion, are works by~\cite{jeong2024multi} and~\cite{zheng2025efficientmultipersonmotionprediction}. However, holistic models often compromise on individual sub-task benchmarks, introducing a gap.  

To close this gap, we present a general and, in hindsight, very simple approach to 3D human motion prediction: \emph{SimpliHuMoN}. SimpliHuMoN is built on a stack of self-attention modules to effectively capture both spatial dependencies within a single pose and temporal relationships across the entire motion sequence. This design enables us to model a wide range of complex motion dynamics while maintaining a streamlined, efficient framework. Our method employs unified, end-to-end training, which improves training stability and overall performance. Our contribution is the empirical finding that unified self-attention over context and queries achieves compelling results across pose, trajectory, and combined-prediction benchmarks using one streamlined architecture.

We validate our approach through extensive experiments on a wide range of public datasets, including Human3.6M \citep{ionescu2013human36m} and AMASS \citep{mahmood2019amass} for pose prediction, ETH-UCY \citep{lerner2007crowds, pellegrini2009youll} and SDD \citep{robicquet2016learning} for trajectory prediction, as well as MOCAP-UMPM \citep{cmu2003mocap, vanderaa2011umpm} and 3DPW \citep{vonmarcard2018recovering} for combined pose and trajectory tasks. Our results show that our model matches or outperforms current best methods across various metrics while being efficient. 

The key contributions of this paper are:
\begin{itemize}[leftmargin=*]
\item We introduce SimpliHuMoN, a simple, unified transformer architecture that handles pose, trajectory, and holistic human motion prediction within a single, end-to-end framework.
\item Our minimalist approach matches or outperforms specialized human motion prediction models.
\end{itemize}

\section{SimpliHuMoN}
\label{method}

\begin{figure*}[t]
    \centering
    \includegraphics[width=\textwidth]{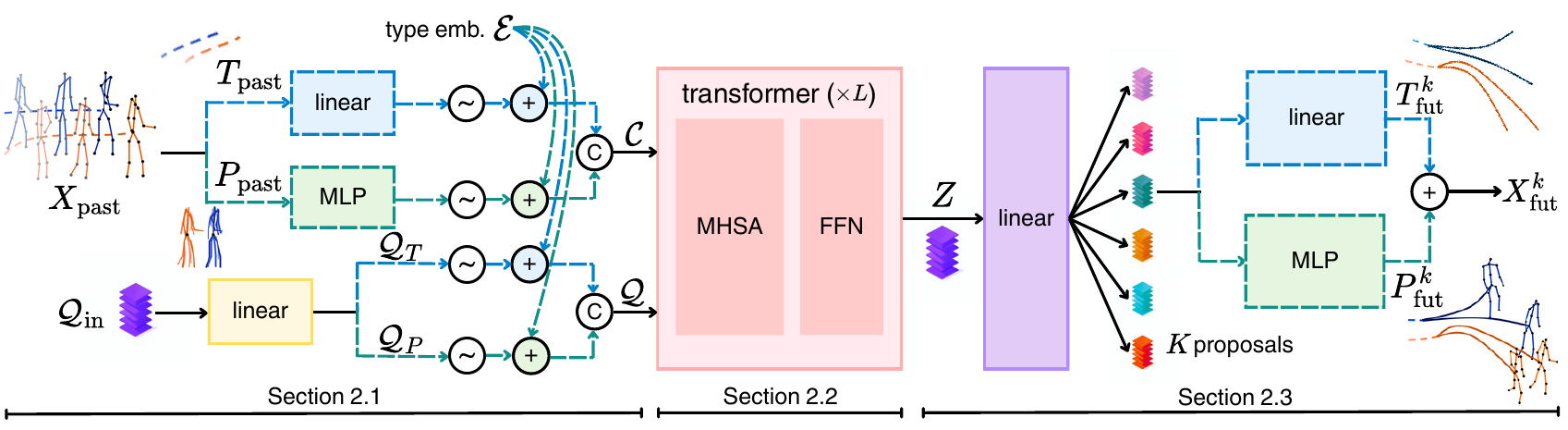}
    \caption{An overview of our architecture. Past observations of 3D poses ($P_{\text{past}}$) and trajectories ($T_{\text{past}}$) are jointly processed by an encoder. Learnable input queries ($\mathcal{Q}_{\text{in}}$), representing potential future states, interact with the encoded past motion within a decoder to produce $K$ distinct future motion proposals ($X_{\text{fut}}^k$) for all agents over a specified horizon.}
    \label{fig:model_arch}
    \vspace*{-4mm}
\end{figure*}

We propose a 3D human motion prediction model based on a transformer decoder architecture. The model is designed to be as elementary as possible, learning a mapping from a person's past movements to their future movements while accommodating various input and output configurations. 

The input $X_\text{past}$ consists of two components, each over a historical time horizon of $H$ timesteps. On the one hand, the trajectory $T_\text{past} \in \mathbb{R}^{H \times 3}$ represents the path of a root joint (\eg, the hip). On the other hand, the relative body pose $P_\text{past} \in \mathbb{R}^{H \times M \times 3}$ represents the state of $M$ joints relative to the root joint. Our model can operate on either of these inputs individually or on both combined: for trajectory prediction, the model only operates on $T_\text{past}$; for pose prediction, the model only operates on $P_\text{past}$; and for joint pose and trajectory prediction, the model operates on both. Each task instantiates the context tokens, queries, type embeddings, and output heads for its available modalities.

The model aims to predict the future state $X_\text{fut}$, over a prediction horizon of $F$ timesteps. To capture the uncertainty of motion, following prior work~\citep{jeong2024multi}, the model generates $K$ distinct proposal states, \ie, $X_\text{fut} = (X_\text{fut}^1, ..., X_\text{fut}^K)$. Each proposal $X_\text{fut}^k$, $k\in\{1, ..., K\}$, consists of a complete predicted future state. 
The composition of $X_\text{fut}^k$ mirrors that of the input; it can include a future root trajectory $T_\text{fut} \in \mathbb{R}^{F \times 3}$, a future relative body pose $P_\text{fut}\in \mathbb{R}^{F \times M \times 3}$, or both, depending on the input. 

\noindent\textbf{Overview of our method.} As illustrated in Figure~\ref{fig:model_arch}, our model begins by independently processing the historical observations $X_\text{past}$ and a set of learnable query tokens $\mathcal{Q}_\text{in} = (\mathcal{Q}_\text{in}^1, ..., \mathcal{Q}_\text{in}^F) \in \mathbb{R}^{F \times 3}$ into a context tensor $\mathcal{C}$ and a query tensor $\mathcal{Q}$  
respectively (Sec.~\ref{enc}). A self-attention-based transformer then processes the tensors (Sec.~\ref{dec}). Finally, a multi-modal prediction head regresses the decoder's output $Z$ into $K$ distinct trajectories and pose hypotheses to give the final output, $X_\text{fut}$ (Sec.~\ref{out}). We describe the training procedure and model configurations in Sec.~\ref{training}.

\subsection{Input Processing and Embedding Module} \label{enc}
This module prepares the input data for the transformer decoder by normalizing it and mapping it into a shared high-dimensional latent space of dimension $d_\text{model}$. It creates two main tensors: a context tensor, $\mathcal{C}$, from past observations and a query tensor, $\mathcal{Q}$, from a set of learnable parameters.

\noindent\textbf{Past Context Encoding.}
To compute the context tensor ${\cal C}$, the historical input sequence is processed using one or both of two parallel streams, depending on the task: one for trajectory $T_\text{past}$ and one for relative body pose $P_\text{past}$.

\noindent\emph{Root Trajectory Processing.} The 3D coordinates of the root joint are extracted from the input sequence. To normalize the motion, the root's position at the final input frame is subtracted from all historical root positions. This normalized trajectory is then projected into the $d_\text{model}$-dimensional embedding space by a linear layer.

\noindent\emph{Relative Pose Processing.} The pose is represented relative to the root (hip) joint for each timestep. If a dataset provides absolute coordinates, we normalize the pose by subtracting the root joint's position from all other body joint positions. This relative pose vector is then processed by a two-layer MLP with GELU activation, which outputs an embedding of dimension $d_\text{model}$.

After their initial embedding, both streams are enhanced. First, a sinusoidal positional encoding is added to each sequence to encode the specific position of each of the $H$ timesteps along the time axis. Then, a learnable type embedding $\mathcal{E}\in\mathbb{R}^{d_\text{model}}$ is added to each token. One vector is learned per modality, broadcast over time, and shared by the past-context and future-query tokens of that modality. Finally, the processed sequences are concatenated (if both are present) along the sequence dimension to form the final context tensor, $\mathcal{C}$. The shape of $\mathcal{C}$ is therefore $\mathbb{R}^{2H \times d_\text{model}}$ for combined inputs, and $\mathbb{R}^{H \times d_\text{model}}$ when only a single input modality is provided.

\noindent\textbf{Future Query Generation.}
The queries used to prompt the decoder are learnable tensors $\mathcal{Q}_\text{in} \in \mathbb{R}^{F \times 3}$. Similar to the object queries in DETR \citep{carion2020detr} or learnable soft prompts in language modeling \citep{lester-etal-2021-power}, these are input-independent parameters optimized during training. They serve as initial ``slots'' for the $F$ future timesteps, providing the decoder with a temporal structure to fill based on the context. These tokens are first projected into the $d_\text{model}$-dimensional space by a linear layer. The resulting sequence is then explicitly split into trajectory $\mathcal{Q}_T\in \mathbb{R}^{ F\times d_\text{model}}$ and pose $\mathcal{Q}_P\in \mathbb{R}^{ F\times d_\text{model}}$ queries if both modalities are required. Similar to the past context encoding, these query sequences are enriched with positional encodings and their corresponding type embeddings. The two query sequences are then concatenated (if both are present) to create the final query tensor, $\mathcal{Q}$ $(\in \mathbb{R}^{2F\times d_\text{model}}$ for combined inputs, $\mathbb{R}^{F\times d_\text{model}}$ for single), ensuring that it mirrors the composition and format of $\mathcal{C}$.

This explicit separation of queries into trajectory and pose streams enables the model's flexibility. The architecture learns a strong association between each query type and its corresponding output modality, reinforced by the type embeddings. This allows the same model to handle different tasks without any architectural modifications. 

\subsection{Transformer Decoder} \label{dec}
The core of SimpliHuMoN is a decoder-only transformer built from multi-head self-attention (MHSA) blocks. We concatenate the context $\mathcal{C}$ and query $\mathcal{Q}$ tensors along the temporal dimension to form a unified input sequence $[\mathcal{C}; \mathcal{Q}] \in \mathbb{R}^{(H+F) \times d_\text{model}}$. MHSA allows every context and query token to attend directly to every other token, enabling bidirectional information flow at every layer. The standard transformer encoder-decoder formulation~\citep{vaswani2017attention} provides the comparison architecture evaluated in App.~\ref{sec:app:addexp}.
For enhanced training stability, we employ pre-LayerNorm with Root Mean Square Layer Normalization (RMSNorm) and a standard Feed-Forward Network (FFN) with GELU activation.

After passing through the stack of $L$ decoder layers, the model produces an output tensor, $Z$, with the exact dimensions as the input query ${\cal Q}$. Having attended to the full context, these output query tokens now serve as rich, context-aware representations, ready to be mapped to future predictions by the output heads. This unified architecture is task-agnostic: whether the input $\mathcal{C}$ contains only trajectory, only pose, or both, the self-attention mechanism naturally adapts to model the available dependencies.

\subsection{Multi-Modal Prediction Heads} \label{out}
To account for the stochastic nature of the prediction task, the prediction head decodes the final latent representation from the decoder into $K$ distinct future hypotheses. A single linear layer projects the decoder's output tensor $Z$ (shape $[F, d_\text{model}]$) to a tensor of shape $[F, K \times C]$, where $C$ is the output dimension (\eg, 3 for trajectory, $M \times 3$ for pose). This is then reshaped to $[F, K, C]$, creating $K$ parallel branches. Two dedicated output heads then process each branch, if both are being modeled, to regress the future root trajectory ($T_\text{fut}^k$) and body pose ($P_\text{fut}^k$), respectively, ensuring each of the $K$ proposals is a complete and comparable hypothesis. Architecturally, these heads mirror the input processing module: a linear layer regresses the trajectory, and a two-layer MLP regresses the pose, effectively inverting the initial embedding process.

\subsection{Implementation Details} \label{training}
The model is trained end-to-end using a ``winner-takes-all'' loss, where gradients are backpropagated only through the single hypothesis $k$ that minimizes the Euclidean distance to the ground truth future. Formally, the training loss $\mathcal{L}$ for a given ground truth $X_\text{fut}^\text{gt}$ is computed via
\begin{equation}
\label{eq:loss}
  \mathcal{L}(X_\text{past},X_\text{fut}^\text{gt}) = \text{min}_{k\in\{1, ..., K\}} \|X_\text{fut}^\text{gt}-X_\text{fut}^k(X_\text{past})\|_2, 
\end{equation}
where $X_\text{fut}^k(X_\text{past})$ is the $k^\text{th}$ prediction hypothesis computed from $X_\text{past}$ via the model, as illustrated in Figure~\ref{fig:model_arch}. For each sample, gradients update the winning hypothesis. Independently initialized slices of the multi-modal projection specialize toward different regions of the future distribution through these sample-specific assignments. This winner-takes-all objective supplies the diversity mechanism used in our experiments.

We report results for two configurations: a ``wide'' model $(L=6, d_\text{model}=192)$ and a ``deep'' model $(L=16, d_\text{model}=48)$. In all experiments, we use the AdamW optimizer $(\beta_1=0.95, \beta_2=0.999)$ with a weight decay of $10^{-4}$. All models are trained for 300 epochs with a batch size of 64 and standard data augmentation on one NVIDIA RTX A6000 GPU. The number of modes, $K$, is set as a hyperparameter. For a fair evaluation, we strictly follow prior work and choose $K$ accordingly.

\section{Experiments}
\label{experiments}
\subsection{Datasets}
We evaluate our model on several standard benchmarks to cover a range of motion prediction tasks. For 3D human pose prediction, we use \colorbox{posecolor}{Human3.6M} \citep{ionescu2013human36m}, a large-scale lab-based dataset, and \colorbox{posecolor}{AMASS} \citep{mahmood2019amass}, a comprehensive motion capture archive used for generative modeling. For trajectory forecasting, we use the pedestrian datasets \colorbox{trajcolor}{ETH-UCY} \citep{lerner2007crowds, pellegrini2009youll} and the \colorbox{trajcolor}{Stanford Drone Dataset (SDD)} \citep{ro2019display}, which contains varied persons from an aerial view. Finally, we evaluate joint pose and trajectory prediction using \colorbox{bothcolor}{Mocap-UMPM} \citep{cmu2003mocap,vanderaa2011umpm}, a mixed dataset of Mocap and UMPM containing synthesized human interaction between three people, and \colorbox{bothcolor}{3DPW} \citep{vonmarcard2018recovering}, a dataset with two people traversing a real-world environment. We report results on each benchmark after training our model on its respective dataset in Table~\ref{tab:my_big_table}, which uses the same color scheme to visually group the results.

\subsection{Metrics}
We evaluate our model following common practice for multi-modal models that generate $K$ proposals, reporting the minimum error among all generated proposals. For pose prediction, we report the minimum Average/Final Displacement Error (ADE/FDE) averaged across all body joints over  \colorbox{posecolor}{$K=7$} proposals, following~\cite{hosseininejad2025motionmap}. For trajectory prediction, we report the ADE/FDE on the root joint over  \colorbox{trajcolor}{$K=20$} proposals, following~\cite{yao2024trajclip}. In the combined pose and trajectory prediction task, we assess local and global accuracy over  \colorbox{bothcolor}{$K=6$} proposals, following~\cite{jeong2024multi}. All methods within each task are evaluated with the same proposal budget. For the combined task, we use two metrics: Aligned mean per-joint Position Error (APE), which measures pose error after root alignment, and Joint Precision Error (JPE), which measures the overall error of all joints in the world coordinate system. Consistent with prior work, for datasets containing multiple people, the final reported metric is the average error across individuals.

\begin{table*}[t]
\centering
\caption{Detailed comparison of model performance. Lower values are better $(\downarrow)$, with the best results shown in \textbf{bold}. An asterisk (*) denotes models we recomputed for this setup, a dagger ($\dagger$) marks models adapted for the specific task, while a ($\wedge$) notes models that use external training data.}
\label{tab:my_big_table}
\small
\setlength{\aboverulesep}{0pt}
\setlength{\belowrulesep}{0pt}

\begin{adjustbox}{width=\linewidth}
\begin{tabular}{ll >{\columncolor{posecolor}}c >{\columncolor{posecolor}}c >{\columncolor{trajcolor}}c >{\columncolor{trajcolor}}c >{\columncolor{bothcolor}}c >{\columncolor{bothcolor}}c} 
\toprule
& & \multicolumn{2}{>{\columncolor{posecolor}}c}{\textbf{Pose Prediction}} 
& \multicolumn{2}{>{\columncolor{trajcolor}}c}{\textbf{Trajectory Prediction}} 
& \multicolumn{2}{>{\columncolor{bothcolor}}c}{\textbf{Pose + Trajectory Prediction}} \\
\midrule
& \thead{Dataset \\ In/Out (s) \\ Metric} 
& \thead{Human3.6M \\ 0.5/2.0 \\ ADE/FDE$\downarrow$} & \thead{\makebox[6.5em][c]{AMASS} \\ 0.5/2.0 \\ ADE/FDE$\downarrow$}
& \thead{ETH-UCY (Avg) \\ 3.2/4.8 \\ ADE/FDE$\downarrow$} & \thead{\makebox[6.5em][c]{SDD} \\ 3.2/4.8 \\ ADE/FDE$\downarrow$}
& \thead{MOCAP-UMPM \\ 1.0/2.0 \\ APE/JPE$\downarrow$} & \thead{\makebox[6.5em][c]{3DPW} \\ 0.8/1.6 \\ APE/JPE$\downarrow$} \\
\midrule

\multirow{6}{*}{\STAB{\rotatebox[origin=c]{90}{Pose}}}
& DivSamp      & 0.48/0.68          & 0.48/0.64          & -- & -- & -- & -- \\
& BeLFusion    & 0.44/0.60          & 0.35/0.48          & -- & -- & -- & -- \\
& CoMusion     & 0.43/0.61          & 0.31/0.46          & -- & -- & -- & -- \\
& Motionmap    & 0.47/0.60          & 0.32/0.45          & -- & -- & -- & -- \\
& SkeletonDiff & 0.64/0.77* & 0.56/0.71* & -- & -- & -- & -- \\
& SLD          & \textbf{0.42}/0.59*& \textbf{0.30}/0.45*& -- & -- & -- & -- \\
\midrule

\multirow{6}{*}{\STAB{\rotatebox[origin=c]{90}{Traj}}}
& MID                & -- & -- & 0.21/0.38 & 7.61/14.32    & -- & -- \\
& GP-Graph           & -- & -- & 0.23/0.39 & 9.10/13.76    & -- & -- \\
& TUTR               & -- & -- & 0.21/0.36 & 7.76/12.69    & -- & -- \\
& SingularTraj.      & -- & -- & 0.22/0.34 & 7.26/12.58    & -- & -- \\
& TrajCLIP           & -- & -- & \textbf{0.18}/0.33$^{\wedge}$ & 6.29/11.79$^{\wedge}$ & -- & -- \\
& NMRF               & -- & -- & 0.19/\textbf{0.32} & 7.20/11.29    & -- & -- \\
\midrule

\multirow{4}{*}{\STAB{\rotatebox[origin=c]{90}{Both}}}
& T2P          & 0.80/1.03$^{\dagger}$ & 0.63/0.94$^{\dagger}$ & 0.19/0.39$^{\dagger}$ & 8.11/8.59$^{\dagger}$  & 151.7/262.7 & 150.0/236.2 \\
& EMPMP        & 0.45/0.72$^{\dagger}$ & 0.42/0.65$^{\dagger}$ & 0.63/0.72$^{\dagger}$ & 10.2/10.5$^{\dagger}$ & 146.5/250.4* & 150.6/235.4* \\
& \cellcolor{highlightgray}Ours (wide)  & \cellcolor{highlightgray}\textbf{0.42}/0.59    & \cellcolor{highlightgray}0.31/\textbf{0.45} & \cellcolor{highlightgray}\textbf{0.18/0.32}    & \cellcolor{highlightgray}6.70/7.63    & \cellcolor{highlightgray}\textbf{125.7}/212.7 & \cellcolor{highlightgray}\textbf{142.9}/\textbf{231.0} \\
& \cellcolor{highlightgray}Ours (deep)  & \cellcolor{highlightgray}0.44/\textbf{0.57}    & \cellcolor{highlightgray}0.35/0.47          & \cellcolor{highlightgray}0.19/\textbf{0.32}    & \cellcolor{highlightgray}\textbf{6.26/7.61}    & \cellcolor{highlightgray}131.4/\textbf{211.8} & \cellcolor{highlightgray}148.9/231.5 \\
\bottomrule
\end{tabular}
\end{adjustbox}
\end{table*}

\subsection{Baselines}
We compare our method against a wide range of state-of-the-art models across three distinct prediction tasks. In the domain of pose-only prediction, we evaluate against several recent generative approaches, including DivSamp \citep{dang2022diverse}, and prominent diffusion-based models such as BeLFusion \citep{barquero2023belfusion}, CoMusion \citep{sun2024comusion}, and SkeletonDiff \citep{curreli2025nonisotropic}. Our comparison in this category also includes Motionmap \citep{hosseininejad2025motionmap} and the state-space diffusion model SLD \citep{xu2024learning}. For trajectory-only prediction, we benchmark against MID \citep{gu2022stochastic}, GP-Graph \citep{bae2022learning}, TUTR \citep{shi2023trajectory}, SingularTrajectory \citep{bae2024singulartrajectory}, vision-language model TrajCLIP \citep{yao2024trajclip}, and NMRF \citep{fang2025neuralized}. Finally, for multi-person motion prediction, which involves forecasting combined human trajectories and poses, we include EMPMP \citep{zheng2025efficientmultipersonmotionprediction} and T2P \citep{jeong2024multi}.

\subsection{Quantitative Results}
Our proposed model demonstrates versatile and robust performance, matching or improving upon state-of-the-art results across a diverse range of motion prediction tasks, as shown in Table~\ref{tab:my_big_table}. Its success as a generalist architecture is particularly noteworthy given that many prior methods  incorporate domain-specific inductive biases. For instance, top-performing baselines often rely on operations such as the Discrete Cosine Transform (DCT) to model motion in the frequency domain \citep{xu2024learning} or employ graph convolutional networks (GCNs) to explicitly encode the body's kinematic structure \citep{sun2024comusion}. The results for our two primary configurations---a ``wide'' model and a ``deep'' model---highlight the effectiveness of our simple, unified approach in challenging established, task-specific methods. 

On the \colorbox{posecolor}{Human3.6M} benchmark, our model's performance matches the leading methods in Average Displacement Error (ADE) while outperforming compared methods in Final Displacement Error (FDE). This strength in long-term prediction is further confirmed on \colorbox{posecolor}{AMASS}, where it again surpasses existing models on the FDE metric. This success illustrates that attention-based transformers can effectively and accurately model high-dimensional pose data. Notably, our model achieves this performance in a single, deterministic forward pass. This differs from the iterative sampling process required for inference by leading generative models \citep{curreli2025nonisotropic, sun2024comusion}.

On trajectory prediction, our ``wide'' model's performance is on par with the current best techniques, matching the leading results on both ADE and FDE metrics for the \colorbox{trajcolor}{ETH-UCY} dataset.
Given that these scenes can contain up to 57 pedestrians, it is worth noting that our straightforward transformer-based architecture can predict successfully in crowded environments. Different from prior work, our transformer architecture does not rely on external knowledge from massive, pre-trained vision-language models, as in TrajCLIP, or on continuous, field-based scene representations used by NMRF. Furthermore, on the \colorbox{trajcolor}{SDD} benchmark, both of our models outperform prior work, with our deep configuration improving on FDE by 32\%.

In the comprehensive task of combined pose and trajectory prediction, the advantages of our unified architecture are most prominent. On both the \colorbox{bothcolor}{MOCAP-UMPM} and \colorbox{bothcolor}{3DPW} datasets, our models outperform prior methods like T2P and EMPMP. These prior approaches often rely on multi-stage pipelines that process localized and global aspects of motion in separate stages \citep{jeong2024multi}. In contrast, by jointly modeling pose and trajectory within a single end-to-end framework, our design more effectively captures the coupled dynamics between local body articulation and global root movement, leading to performance gains across all metrics. For instance, on MOCAP-UMPM, our models lower the APE by 10.3\% and the JPE by 15\%.

\begin{wraptable}{r}{0.57\textwidth}
    \vspace{-4mm}
    \centering
    \caption{Throughput mean $\pm$ std calculated over 10 runs on MOCAP-UMPM. All models are run on a NVIDIA RTX A6000 GPU with batch size 64. Higher values are better $(\uparrow)$. A * denotes models we recomputed for this experiment.}
    \label{tab:throughput}
    \small
    \renewcommand\theadfont{\bfseries}
    \centering
    \begin{tabular}{lcc}
    \toprule
    \thead{Model} & \thead{Training Throughput \\ (samples/sec) $\uparrow$} & \thead{Test Throughput\\ (samples/sec) $\uparrow$} \\
    \midrule
    T2P* & 187 $\pm$ 22  & 401  $\pm$ 64     \\
    EMPMP* & 812 $\pm$ 58  & 2041 $\pm$ 129  \\
    Ours (wide)    & 862 $\pm$ 43  & 2251 $\pm$ 140  \\
    Ours (deep)    & \textbf{928} $\pm$ 45 & \textbf{3673} $\pm$ 161\\
    \bottomrule
    \end{tabular}
    \vspace{-4mm}
\end{wraptable}
Additionally, we want to note that our model is computationally very efficient. To demonstrate this, we benchmarked all models that perform joint pose and trajectory prediction on the MOCAP-UMPM dataset, comparing the average number of samples processed per second. Our ``deep'' configuration is not only more accurate but also more computationally efficient than the lightweight EMPMP model, achieving a 14.3\% increase in training throughput and processing test samples at around 1.8 times the speed. Please see Table~\ref{tab:throughput} for details.

\begin{figure*}[t]
    \centering
    \includegraphics[width=\textwidth]{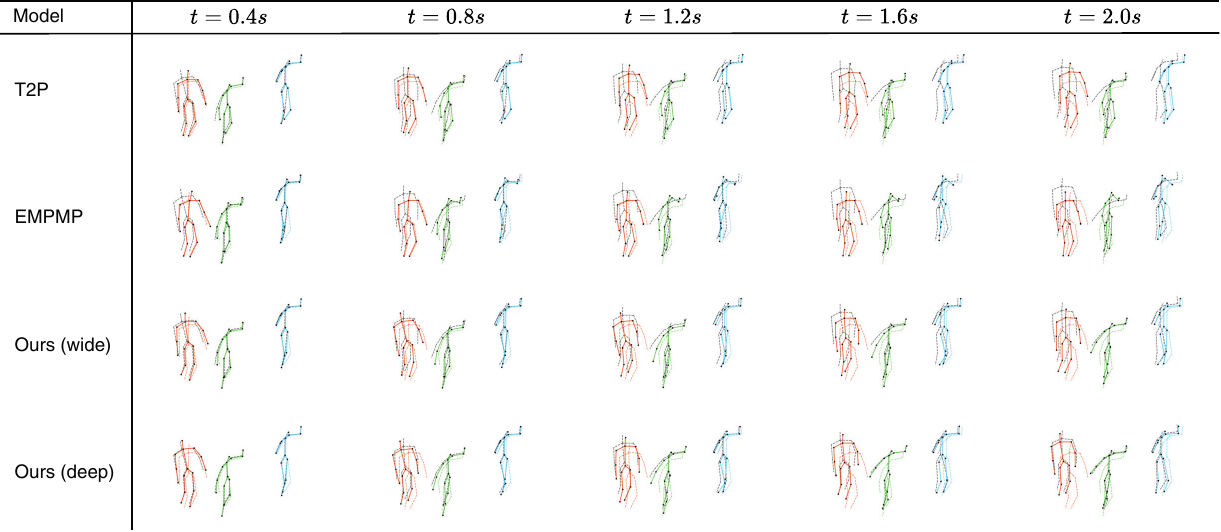}
    \caption{Visualization of predictions on a MOCAP-UMPM scene. Model predictions are in color, and ground truth future poses are black dashes. The last-known input positions are colored dashes.}
    \label{fig:mocap_viz}
    \vspace{-4mm}
\end{figure*}

\subsection{Qualitative Results}
We compare predicted motions on the MOCAP-UMPM dataset in Figure \ref{fig:mocap_viz}, focusing on a challenging scenario where three individuals perform coordinated backward walking. Our ``wide'' and ``deep'' configurations both generate fluid, physically plausible sequences that accurately recover underlying dynamics. Notably, the articulation of the limbs and torso remains realistic across the horizon, demonstrating that SimpliHuMoN learns natural motion manifolds without being constrained by explicit structural or kinematic priors. Our ``deep'' model, in particular, maintains high-fidelity dynamic predictions even at final $t=2.0s$, where temporal drift typically degrades accuracy. 

The limitations of specialized baselines further highlight the advantages of our unified architecture. When faced with high-uncertainty motion, T2P adopts an overly conservative strategy; its predictions quickly collapse toward a static mean pose, sacrificing dynamics to minimize error. Conversely, while EMPMP seeks dynamism, it struggles with physical plausibility. Its outputs exhibit artifacts, such as unnatural arm postures (green agent) and unnatural, disjointed leg movements (blue agent). These results underscore that SimpliHuMoN not only achieves superior quantitative scores but also produces more coherent and realistic motion than multi-stage or fragmented methods.

\begin{figure*}[t]
    \centering
    \includegraphics[width=\linewidth]{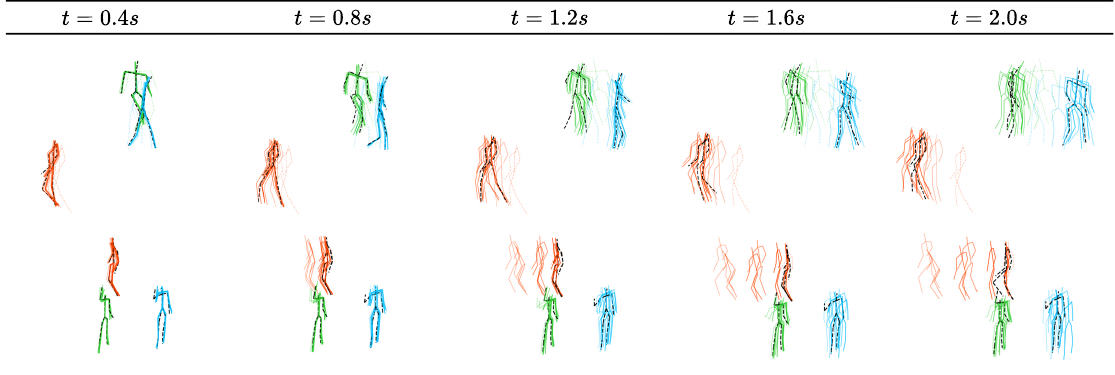}
    \caption{Visualization of motion proposals ($K=6$) of our (wide) model on MOCAP-UMPM data. All model predictions are in color. Ground-truth future poses are represented by black dashes, and the last-known input positions are colored dashes.}
    \label{fig:diversity_viz_alt}
    \vspace*{-4mm}
\end{figure*}

Beyond accuracy, we evaluate the model's ability to capture the stochastic nature of human behavior. Figure \ref{fig:diversity_viz_alt} visualizes the distribution of all $K=6$ proposals generated by our ``wide'' model. Rather than producing minor variations of a single path, the model captures distinct, high-level intent. For \eg, the red agent is simultaneously predicted to walk straight, decelerate to a stop, or execute a turn. Each of the $K$ hypotheses represents a physically coherent and plausible future, highlighting that our method effectively maps the multi-modal distribution of human motion.

\subsection{Ablation Studies}
\label{ablation}
In this section, we conduct ablation studies to investigate the impact of our model's key components and hyperparameters. We perform these experiments on the MOCAP-UMPM dataset for the joint pose and trajectory prediction task to analyze the effectiveness of our multi-modal prediction head and the trade-offs in our transformer architecture.

\begin{table}[t]
    \centering
    \caption{Comparison of our model's performance with different hyperparameter configurations.}
    \label{tab:ablation_6heads}
    \small
    \begin{minipage}[t]{0.62\textwidth}
        \centering
        \begin{tabular}[t]{ccccc} 
        \toprule
        \textbf{Depth} & \textbf{Embed dim} & \textbf{Total Params} & \textbf{APE} & \textbf{JPE} \\
        \midrule
        8  & 192 & 5.2M & 126.05 & 212.84\\
        6  & 192 & 4.0M & \textbf{125.70} & 212.72\\
        4  & 192 & 2.8M & 126.22 & 213.40 \\
        12 & 96  & 1.9M & 128.47 & 212.08 \\
        6  & 96  & 1.0M & 128.52 & 212.45\\
        12 & 64  & 860K & 130.72 & 212.30 \\
        16 & 48  & 642K & 131.41 & \textbf{211.76}\\
        12 & 48  & 490K & 131.09 & 212.73\\
        16 & 36  & 367K & 134.52 & 215.36 \\
        \bottomrule
        \end{tabular}
    \end{minipage}
    \hfill
    \begin{minipage}[t]{0.35\textwidth}
        \centering
        \vspace{0pt}
        \includegraphics[height=4.3cm]{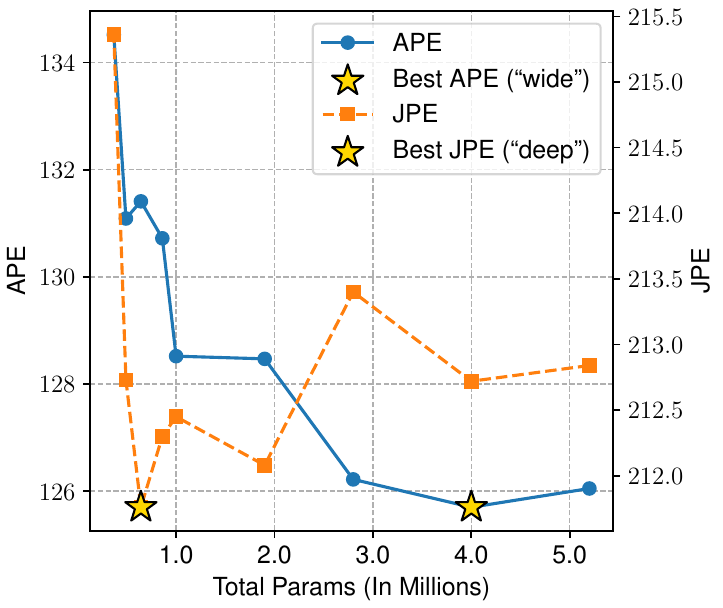}
    \end{minipage}
    \vspace*{-5mm}
\end{table}

\noindent\textbf{Choice of Transformer Hyperparameters.}
Our model's major computations are performed using a simple transformer decoder. We analyze the trade-offs between its depth (number of layers, $L$) and width (embedding dimension, $d_\text{model}$). We experimented with various configurations, keeping the overall parameter count relatively low, to find effective deep net architecture designs (Table~\ref{tab:ablation_6heads}).

The analysis reveals a relationship between depth, width, and predictive accuracy. Our ``wide'' configuration $(L=6, d_\text{model}=192)$ achieves the best APE, suggesting that a more expansive embedding space is beneficial for capturing fine-grained pose details. Decreasing the depth to $L=4$ or increasing it to $L=8$ with the same width leads to a decline in performance, indicating a sweet spot for this configuration.

Conversely, our ``deep'' model $(L=16, d_\text{model}=48)$ obtains the lowest JPE, demonstrating that a deeper stack of attention layers is more effective at modeling complex, long-range spatio-temporal dependencies for global trajectory prediction, even with a constrained embedding dimension. As expected, performance degrades significantly with shallower or narrower architectures. These results validate our choice of ``wide'' and ``deep'' models, as they represent two distinct and highly effective points in the architecture design space, tailored for different aspects of motion prediction.

\noindent\textbf{Effect of Multi-Modal Prediction.}
While multi-modal prediction is standard in trajectory forecasting, state-of-the-art methods for joint pose and trajectory prediction, such as EMPMP, have often favored a deterministic approach, predicting a single future outcome. However, human motion is inherently stochastic, and a single prediction can struggle to capture the full range of plausible futures. We therefore conduct an ablation to explicitly quantify the advantage of our multi-modal prediction head. We compare our model's performance when generating multiple proposals $(K=6)$ against a deterministic setting $(K=1)$, mirroring the setup of prior work \citep{jeong2024multi}.

\begin{wraptable}{r}{0.52\textwidth}
\vspace{-4mm} 
\centering
\setlength{\belowcaptionskip}{2pt}
\caption{Model performance with 2 different modes on MOCAP-UMPM data. Lower values are better $(\downarrow)$.}
\label{tab:k_comparison}
\small
\begin{tabular}{lcccc}
\toprule
& \multicolumn{2}{c}{$K=1$} & \multicolumn{2}{c}{$K=6$} \\
\cmidrule(lr){2-3}\cmidrule(lr){4-5}
\textbf{Model} & \textbf{APE} & \textbf{JPE} & \textbf{APE} & \textbf{JPE} \\
\midrule
T2P            & 154.4           & 366.4          & 151.7           & 262.7          \\
EMPMP          & 147.2           & 283.1          & 146.5           & 250.4          \\
Ours (wide)    & \textbf{145.84} & \textbf{280.8} & \textbf{125.70} & 212.72         \\
Ours (deep)    & 149.35          & 286.96         & 131.41          & \textbf{211.76} \\
\bottomrule
\end{tabular}
\vspace*{-2mm}
\end{wraptable}

The results in Table~\ref{tab:k_comparison} clearly demonstrate the limitations of a deterministic approach. Even in a deterministic setting, our ``wide'' model is already competitive with EMPMP. However, by embracing multi-modality, our model achieves a significant performance gain. The APE improves by 13.8\% and the JPE by a substantial 24.2\%. This highlights that our model doesn't just produce a better single guess; it effectively captures the ambiguity in high-quality future motions. Interestingly, prior works do not benefit from multiple modes to the same degree. For instance, EMPMP's APE barely improves, suggesting its architecture may struggle to generate genuinely distinct futures. While a full analysis of why the baselines are less suited to multi-modal prediction is beyond the scope of this paper, it suggests that our unified architecture is effective at producing a diverse and plausible set of outcomes. %

\noindent\textbf{Benefit of Joint Modeling.}
\begin{table}[t]
    \centering
    \caption{Comparison of the influence of combined pose and trajectory information vs. individual performance on MOCAP-UMPM data. Lower values are better $(\downarrow)$, with the best results shown in \textbf{bold}. A dagger ($\dagger$) marks models adapted for the specific task.}
    \label{tab:train_settings}
    \small
    \setlength{\aboverulesep}{0pt}
    \setlength{\belowrulesep}{0pt}
    \renewcommand\theadalign{cc}
    \renewcommand\theadfont{\bfseries} 
    \begin{tabular}{l
                    >{\columncolor{posecolor}}c
                    >{\columncolor{posecolor}}c
                    >{\columncolor{trajcolor}}c
                    >{\columncolor{trajcolor}}c}
    \toprule
    & \multicolumn{2}{>{\columncolor{posecolor}}c}{\textbf{Pose Prediction}} 
    & \multicolumn{2}{>{\columncolor{trajcolor}}c}{\textbf{Trajectory Prediction}} \\
    \cmidrule(lr){2-5} 
    \thead{Training Method \\ Metric} &
    \thead{Only Pose \\ ADE$\downarrow$/FDE$\downarrow$} & 
    \thead{\makebox[6.5em][c]{Pose + Traj} \\ ADE$\downarrow$/FDE$\downarrow$} &
    \thead{Only Traj\\ ADE$\downarrow$/FDE$\downarrow$} & 
    \thead{\makebox[6.5em][c]{Pose + Traj} \\ ADE$\downarrow$/FDE$\downarrow$} \\ 
    \midrule   
    
    T2P$^\dagger$      & - & 0.61/0.95     & - & 0.20/0.29 \\
    EMPMP$^\dagger$     & - & 0.53/0.65     & - & 0.14/0.20 \\
    \cellcolor{highlightgray}Ours (wide)  & \cellcolor{highlightgray}0.46/0.58  & \cellcolor{highlightgray}\textbf{0.41}/\textbf{0.51} & \cellcolor{highlightgray}0.09/0.18  & \cellcolor{highlightgray}\textbf{0.08/0.16} \\
    \cellcolor{highlightgray}Ours (deep)  & \cellcolor{highlightgray}0.46/0.59    & \cellcolor{highlightgray}0.42/\textbf{0.51}    & \cellcolor{highlightgray}0.09/0.18           & \cellcolor{highlightgray}\textbf{0.08}/0.17   \\
    \midrule
    Improvement (\%) & \cellcolor{white} & \cellcolor{white}\textcolor{green}{-11.4/-12.2} & \cellcolor{white} & \cellcolor{white}\textcolor{green}{-13.7/-11.9}\\
    \bottomrule
    \end{tabular}
    \vspace{-7mm}
\end{table}
A core hypothesis of our work is that pose and trajectory are deeply intertwined, and that jointly modeling them improves predictions of both. To test this, we trained ``pose-only'' and ``trajectory-only'' variants of SimpliHuMoN and compared them to the full model. 
As shown in Table \ref{tab:train_settings}, pose prediction accuracy improves by $\sim$11-12\% when trajectory information is provided as an input, compared to a ``pose-only'' variant. Conversely, trajectory prediction accuracy improves by $\sim$12-14\%  when pose data is included. This provides quantitative evidence that our unified framework learns the physical coupling between pose and trajectory, using information from one to refine its predictions for the other.

\section{Related Work}
\label{related_work}
Human motion requires a holistic assessment, as local body articulation (pose) and global displacement (trajectory) are deeply intertwined. Our research community, however, has largely tackled motion prediction by decomposing this process into specialized sub-problems: pose and trajectory prediction. 
This specialization has driven progress on narrow benchmarks but created a dichotomy: specialized models struggle to generalize, while holistic models struggle to compete on established task-specific leaderboards. This ``benchmark effect'' has incentivized architectural innovations with increasingly elaborate models gaining an edge.
In this context, the transformer has emerged as a powerful tool for sequence modeling. However, its application to human motion merely serves as a backbone for other domain-specific modules. This paper challenges that approach. We posit that the transformer's true power lies not in its ability to support additional components but in its inherent capacity to address the problem directly and uniformly.

\subsection{Human Pose Prediction}
The task of human pose prediction involves forecasting a future sequence of 3D skeletal joint locations relative to the root joint based on an observed history of poses \citep[see Appendix C]{hosseininejad2025motionmap}. To address the stochastic nature of human behavior, the field has shifted from deterministic models \citep{GraberCVPR2020, GraberCVPR2020W, medjaouri2022hrstan, ma2022progressively} to generative frameworks, particularly diffusion models. This pursuit of generative fidelity has increased architectural complexity, with methods like BeLFusion \citep{barquero2023belfusion} introducing a ``behavioral latent space'' and CoMusion \citep{sun2024comusion} employing a hybrid Transformer-GCN architecture that operates in the Discrete Cosine Transform (DCT) \citep{mao2021generating} space to model skeletal kinematics explicitly. Recent methods like SkeletonDiff \citep{curreli2025nonisotropic} and SLD \citep{xu2024learning} focus on skeleton-aware generation or long-sequence efficiency, while non-diffusion approaches like MotionMap \citep{hosseininejad2025motionmap} introduce novelties such as multi-stage heatmap pipelines. 

\subsection{Human Trajectory Prediction}
Trajectory forecasting aims to predict an agent's root joint trajectory, a task complicated by latent intent, social interactions, and environmental constraints. Recent state-of-the-art approaches have often relied on massive external knowledge sources or engineered, multi-stage pipelines. A prominent trend involves leveraging large foundation models; TrajCLIP \citep{yao2024trajclip}, for example, incorporates knowledge from vision-language models (VLMs) to provide contextual cues, effectively outsourcing the learning problem. Another approach involves building frameworks for generality, such as Singular Trajectory \citep{bae2024singulartrajectory}, whose ``universal'' status results from a pipeline that combines Singular Value Decomposition with a diffusion-based refiner, or NMRF \citep{fang2025neuralized}, which uses modules such as continuous, field-based scene representations.

\subsection{Combined Pose and Trajectory Prediction}
The simultaneous prediction of pose and trajectory is where the limitations of specialized architectures become most apparent. This task requires modeling the critical coupling between local and global movement. Early work such as Tripod \citep{Adeli2021TRiPODHT} and work by~\cite{Zaier_2023_ICCV} established the importance of forecasting these dynamics jointly, typically employing graph-based or multi-branch architectures to capture the dependencies. More recent approaches have explored pre-training strategies, such as Multi-transmotion \citep{gao2024multi}, to learn generalizable motion representations. Despite this progress, prior work has typically imposed strong architectural priors on how pose and trajectory information should interact. T2P \citep{jeong2024multi} employs a sequential, ``coarse-to-fine'' strategy, first predicting the global trajectory and then conditioning the pose prediction on that result. This design imposes a one-way causal assumption that the trajectory dictates the pose and is susceptible to error propagation. An alternative, seen in EMPMP \citep{zheng2025efficientmultipersonmotionprediction}, uses parallel branches to process local and global information separately before fusion. This avoids direct error propagation but imposes a prior that local and global features are separable. This rigid separation may preclude the model from learning  deeply intertwined representations where local and global dynamics are jointly encoded from the outset. Consequently, although EMPMP was explicitly designed to be ``lightweight'', its architecture is built from individually light but intricately integrated components and struggles to leverage hardware parallelism effectively.

\section{Conclusion}
\label{conclusion}
This paper introduces SimpliHuMoN, a streamlined and unified transformer-based model. The model addresses the prevailing trend of specialization in human motion prediction. We demonstrate that a single, end-to-end framework effectively learns the dynamics of human movement across various tasks. Extensive experiments across a wide range of standard benchmarks validate this approach, showing that the model achieves state-of-the-art accuracy while also proving more computationally efficient than prior methods. Ultimately, this work provides evidence that well-designed, minimalist foundations can achieve compelling and generalizable motion prediction. We hope to encourage further study of streamlined architectures for this problem.

\paragraph{Limitations.} Our model predicts people independently, which limits its treatment of crowded or tightly coupled scenes. Neighbor histories could be incorporated as additional typed context tokens in future work. The current state representation uses joint positions, but since the model backbone accepts vector-valued frame inputs, future work could extend it to SMPL pose and shape parameters with mesh-specific training and evaluation. Finally, best-of-$K$ evaluation measures proposal coverage so single-future applications would benefit from confidence scores or a hypothesis-ranking mechanism.

\noindent\textbf{Acknowledgements.} Work supported in part by NSF grants 2008387, 2045586, 2106825, 2551476, MRI 1725729, and NIFA award 2020-67021-32799.

\newpage
\bibliographystyle{plainnat}
\bibliography{main}

\newpage
\appendix

\section*{Appendix: SimpliHuMoN: Simplifying Human Motion Prediction}

\noindent This appendix is structured as follows: In Sec.~\ref{sec:app:adddata} we provide training, dataset and metric details. 
In Sec.~\ref{sec:app:addexp} we detail additional experimental results. 
In Sec.~\ref{sec:app:joint} we discuss joint training results, detailing the methodology for creating a universal motion model across heterogeneous datasets.  
In Sec.~\ref{sec:app:diversity} we evaluate prediction diversity through mode utilization analysis and comparison with generative baselines.
In Sec.~\ref{sec:app:qual} we present additional qualitative visualizations and an analysis of model failure cases.
Finally, in Sec.~\ref{sec:app:web} we highlight the website which is part of the supplementary material.

\section{Additional Implementation Details}
\label{sec:app:adddata}
\subsection{Architecture and Training Hyperparameters}
Table~\ref{tab:implementation_details} consolidates the settings shared across pose-only, trajectory-only, and combined prediction. Only the proposal budget $K$ changes with the benchmark protocol.

\begin{table}[h]
\centering
\caption{Model settings used in all experiments.}
\label{tab:implementation_details}
\small
\begin{tabular}{ll}
\toprule
\textbf{Component} & \textbf{Setting} \\
\midrule
Decoder layers / $d_\text{model}$ & 6 / 192 (wide); 16 / 48 (deep) \\
Attention heads & 4 \\
FFN hidden dimension / activation & $2d_\text{model}$ / GELU \\
Normalization & Pre-norm RMSNorm \\
Positional encoding & Sinusoidal, on context and query streams \\
Type embedding & $\mathbb{R}^{d_\text{model}}$ per modality; $10^{-3}\!\times\mathcal{N}(0,1)$ initialization \\
Trajectory embedding / head & Linear / linear \\
Pose embedding / head & Two-layer MLP / two-layer MLP, with GELU \\
Multi-modal projection & Linear $[F,d_\text{model}]\rightarrow[F,K C]$, reshaped to $[F,K,C]$ \\
\bottomrule
\end{tabular}
\end{table}

\subsection{Sources and Processing of Data}
All experiments are conducted on publicly available, open-source datasets. To ensure a fair and direct comparison with prior work, we strictly adhere to the established data-processing and evaluation protocols of recent top-performing methods for each prediction task. 
The specific protocols are as follows:
For pose prediction on the Human3.6M and AMASS datasets, we follow the data processing methodology, sequence lengths, and evaluation splits established by BeLFusion \citep{barquero2023belfusion}.
For trajectory prediction on the ETH-UCY and SDD, our data handling and evaluation procedures align with the protocol set forth by NMRF \citep{fang2025neuralized}.
For combined pose and trajectory prediction on the MOCAP-UMPM and 3DPW datasets, we adopt the data preparation and processing pipeline outlined by T2P \citep{jeong2024multi}.

\subsection{Metric Formulae}
Given the predicted motion proposal $X_\text{fut}^k = \{x_{t,m}^{k}\} \in \mathbb{R}^{F\times M \times 3}$ for $k \in \{1, 2, ..., K\}$ across $F$ time frames with $M$ joints per person, along with the corresponding ground truth $X_\text{fut}^\text{gt}= \{x_{t,m}^\text{gt}\}$, the following metrics are used for evaluation. For multi-modal predictions, we follow common practice and report the minimum error among all $K$ generated proposals for each metric (\eg, minADE, minFDE). Consistent with prior work, for datasets containing multiple people, the final reported error is the average of the metric computed for all individuals. All metrics in the main paper are reported for the final output timestep, $t=F$.

\paragraph{APE.} Aligned mean per joint Position Error (APE) is used as a metric to evaluate the forecasted local motion. Euclidean distance (in mm) of each joint relative to the root joint is averaged over all joints for a given timestep, $t$:
\begin{equation}
\text{APE}_t(X_\text{fut}^\text{gt},X_\text{fut}^k) = \frac{1}{M} \sum_{m=1}^{M} \lVert(x_{t,m}^\text{gt}-x_{t,\text{hip}}^\text{gt})-(x_{t,m}^k-x_{t,\text{hip}}^k)\rVert_2.
\end{equation}

\paragraph{JPE.} Joint Precision Error (JPE) evaluates both global and local predictions by the average Euclidean distance (in mm) of all joints for a given timestep, $t$:
\begin{equation}
\text{JPE}_t(X_\text{fut}^\text{gt},X_\text{fut}^k) = \frac{1}{M} \sum_{m=1}^{M} \lVert x_{t,m}^\text{gt}-x_{t,m}^k\rVert_2.
\end{equation}

\paragraph{ADE.} Average Displacement Error (ADE) measures the Euclidean distance (in m) between the ground truth and predicted sequences, averaged over all joints and all future time frames:
\begin{equation}
\text{ADE}(X_\text{fut}^\text{gt},X_\text{fut}^k) = \frac{1}{F \times M} \sum_{t=1}^{F} \sum_{m=1}^{M} \lVert x_{t,m}^\text{gt}-x_{t,m}^k\rVert_2.
\end{equation}

\paragraph{FDE.} Final Displacement Error (FDE) measures the Euclidean distance (in m) between the ground truth and the prediction, averaged over all joints for a given timestep, $t$:
\begin{equation}
\text{FDE}_t(X_\text{fut}^\text{gt},X_\text{fut}^k) = \frac{1}{M} \sum_{m=1}^{M} \lVert x_{t,m}^\text{gt}-x_{t,m}^k\rVert_2.
\end{equation}

\subsection{Compute} 
The RTX A6000 used for our experiments has 48GB of VRAM; our models consume approximately 2-10 GB during training. Training a single model on any dataset takes ~0.5(3DPW, SDD)- 4 (Human3.6M, AMASS) hours, depending on the dataset size. The full set of experiments reported in this paper, including ablation studies and joint training, required approximately 25 GPU-hours.

\section{Additional Experimental Results}
\label{sec:app:addexp}
\subsection{Per-dataset split on ETH-UCY}
\begin{table}[t]
    \centering
    \caption{Trajectory prediction performance (ADE/FDE) on ETH-UCY. Lower values are better, with the best results shown in \textbf{bold}. A dagger ($\dagger$) marks models adapted for the specific task, while a ($\wedge$) notes models that use external training data.}
    \label{tab:trajectory_results}
    \small
    \setlength{\aboverulesep}{0pt}
    \setlength{\belowrulesep}{0pt}
    \begin{tabular}{l >{\columncolor{trajcolor}}c >{\columncolor{trajcolor}}c >{\columncolor{trajcolor}}c >{\columncolor{trajcolor}}c >{\columncolor{trajcolor}}c>{\columncolor{trajcolor}}c} 
    \toprule
    \textbf{Model} & \textbf{ETH} & \textbf{HOTEL} & \textbf{UNIV} & \textbf{ZARA1} & \textbf{ZARA2} & \textbf{AVG} \\
    \midrule
    MID & 0.39/0.66 & 0.13/0.22 & 0.22/0.45 & 0.17/0.30 & 0.13/0.27 & 0.21/0.38 \\
    GP-Graph & 0.43/0.63 & 0.18/0.30 & 0.24/0.42 & 0.17/0.31 & 0.15/0.29 & 0.23/0.39 \\
    TUTR & 0.40/0.61 & 0.11/0.18 & 0.23/0.42 & 0.18/0.34 & 0.13/0.25 & 0.21/0.36 \\
    SingularTraj. & 0.35/0.42 & 0.13/0.19 & 0.25/0.44 & 0.19/0.32 & 0.15/0.25 & 0.22/0.34 \\
    TrajCLIP$^{\wedge}$ & 0.36/0.57 & \textbf{0.10/0.17} & \textbf{0.19/0.41} & \textbf{0.16/0.28} & \textbf{0.11/0.20} & \textbf{0.18}/0.33 \\
    NMRF & \textbf{0.26/0.37} & 0.11/0.17 & 0.28/0.49 & 0.17/0.30 & 0.14/0.25 & 0.19/\textbf{0.32} \\
    T2P$^{\dagger}$& 0.29/0.55 & 0.15/0.27 & 0.25/0.53 & \textbf{0.16}/0.33 & 0.12/0.26 & 0.19/0.39 \\
    EMPMP$^{\dagger}$ & 0.99/0.98 & 0.70/0.87 & 0.69/0.89 & 0.43/0.50 & 0.32/0.35 & 0.63/0.72 \\
    \rowcolor{gray!20} Ours (wide) & 0.28/0.44 & 0.13/0.24 & 0.24/0.44 & \textbf{0.16}/0.29 & \textbf{0.11}/0.21 & \textbf{0.18/0.32} \\
    \rowcolor{gray!20} Ours (deep) & 0.29/0.44 & 0.14/0.24 & 0.24/0.43 & 0.17/0.29 & 0.13/0.21 & 0.19/\textbf{0.32} \\
    \bottomrule
    \end{tabular}
\end{table}
On the ETH-UCY datasets, our model demonstrates highly competitive performance against leading methods, as detailed in Table~\ref{tab:trajectory_results}. While models like TrajCLIP~\citep{yao2024trajclip} and NMRF~\citep{fang2025neuralized} achieve the best results on some of the individual scenes, our ``wide'' configuration achieves the best overall performance, tying for the best average ADE (0.18) and the best average FDE (0.32).
This result is particularly noteworthy when considering the architectural differences between our model and methods like TrajCLIP. TrajCLIP's strong performance stems from its use of a large, pre-trained VLM to provide rich semantic priors. Specifically, it uses natural-language prompts (\eg, ``a person walking'') to generate contextual embeddings from the VLM's text encoder, which are then fused with visual features to guide trajectory prediction. This approach effectively outsources a part of the learning problem to a large external knowledge base. While powerful, this creates a dependency on computationally heavy external models and assumes that general web-scale knowledge is optimally suited for the fine-grained physics of trajectory prediction.

Our model, in contrast, is entirely self-contained, learning all necessary dynamics exclusively from the provided motion data. The performance difference on the ETH scene, where our model significantly outperforms TrajCLIP, suggests a key advantage of this self-sufficient approach. The ETH dataset represents a scenario where the visual-semantic cues that TrajCLIP relies on are less informative and reliable than in other scenes. In such cases, our model's ability to learn robustly from the motion dynamics alone allows it to generalize more effectively, leading to a more consistent performance profile across all five datasets. This consistency is what enables our model to achieve better average performance without relying on external priors, challenging the notion that they are a prerequisite for top-tier trajectory forecasting.

Furthermore, a key architectural difference is TrajCLIP's explicit modeling of social and environmental interactions through two dedicated modules. They are designed to capture the dynamics among different agents and integrate visual context from the environment to make predictions that are physically consistent with the static scene. The fact that our simpler, non-interactive approach still achieves state-of-the-art average performance highlights the remarkable strength and efficiency of its core motion representation. This also points to a promising avenue for future work: integrating a lightweight interaction mechanism into our powerful architecture could further improve performance.

\subsection{Detailed Metrics Across Key Frames}

\begin{table}[t]
    \centering
    \caption{Comparison of APE/JPE metrics across models and datasets. Lower values are better $(\downarrow)$, with the best results shown in \textbf{bold}. An asterisk (*) denotes models we recomputed for this setup.}
    \label{tab:motion_eval}
    \small
    \resizebox{\textwidth}{!}{%
    \setlength{\aboverulesep}{0pt}
    \setlength{\belowrulesep}{0pt}
    \begin{tabular}{ll >{\columncolor{bothcolor}}c >{\columncolor{bothcolor}}c >{\columncolor{bothcolor}}c >{\columncolor{bothcolor}}c >{\columncolor{bothcolor}}c >{\columncolor{bothcolor}}c >{\columncolor{bothcolor}}c >{\columncolor{bothcolor}}c >{\columncolor{bothcolor}}c}
    \toprule
    & & \multicolumn{5}{>{\columncolor{bothcolor}}c}{\textbf{MOCAP-UMPM}} & \multicolumn{4}{>{\columncolor{bothcolor}}c}{\textbf{3DPW}} \\
    \cmidrule(lr){3-7} \cmidrule(lr){8-11}
    & \textbf{In/Out Length (s)} & \textbf{0.4s} & \textbf{0.8s} & \textbf{1.2s} & \textbf{1.6s} & \textbf{2.0s}  & \textbf{0.4s} & \textbf{0.8s} & \textbf{1.2s} & \textbf{1.6s} \\
    \midrule

    \multirow{4}{*}{\STAB{\rotatebox[origin=c]{90}{\textbf{APE}}}}
    & T2P*       & 71.7 & 107.8 & 120.4 & 137.1 & 151.7 & 98.2 & 114.6 & 135.3 & 150.0 \\
    & EMPMP*     & 60.1 & 96.0 & 116.9 & 131.6 & 146.5 & 96.3 & 111.9 & 134.4 & 150.6 \\
    & \cellcolor{highlightgray}Ours (wide)& \cellcolor{highlightgray} \textbf{57.3} &\cellcolor{highlightgray} \textbf{87.7} & \cellcolor{highlightgray}\textbf{104.5} & \cellcolor{highlightgray} \textbf{115.3} & \cellcolor{highlightgray}\textbf{125.7} & \cellcolor{highlightgray} 92.8 & \cellcolor{highlightgray}\textbf{107.1} & \cellcolor{highlightgray} \textbf{130.0} & \cellcolor{highlightgray}\textbf{142.9} \\
    & \cellcolor{highlightgray}Ours (deep)& \cellcolor{highlightgray} 62.3 & \cellcolor{highlightgray} 89.5 & \cellcolor{highlightgray}107.3 & \cellcolor{highlightgray} 119.0 & \cellcolor{highlightgray} 128.5 &\cellcolor{highlightgray} \textbf{93.2} & \cellcolor{highlightgray} 108.4 & \cellcolor{highlightgray} 131.5 & \cellcolor{highlightgray} 148.9 \\
    \midrule

    \multirow{4}{*}{\STAB{\rotatebox[origin=c]{90}{\textbf{JPE}}}}
    & T2P*       & 70.2 & 139.2 & 160.1 & 226.4 & 262.7 & 107.7 & 142.6 & 181.0 & 236.2 \\
    & EMPMP*     & 68.0 & 123.9 & 170.3 & 219.1 & 250.4 & 103.6 & 140.2 & 179.8 & 235.4 \\
    & \cellcolor{highlightgray}Ours (wide)& \cellcolor{highlightgray} \textbf{64.6} &\cellcolor{highlightgray} \textbf{108.6} & \cellcolor{highlightgray}\textbf{143.9} & \cellcolor{highlightgray} 177.7 & \cellcolor{highlightgray} 212.7 & \cellcolor{highlightgray} \textbf{99.4} & \cellcolor{highlightgray}\textbf{137.3} & \cellcolor{highlightgray} 172.1 &\cellcolor{highlightgray}\textbf{231.0} \\
    & \cellcolor{highlightgray}Ours (deep)& \cellcolor{highlightgray} 68.9 &\cellcolor{highlightgray} 109.9 & \cellcolor{highlightgray} 145.3 & \cellcolor{highlightgray} \textbf{177.2} &\cellcolor{highlightgray}\textbf{210.3} &\cellcolor{highlightgray} 100.1 &\cellcolor{highlightgray} 138.2 & \cellcolor{highlightgray} \textbf{171.6} & \cellcolor{highlightgray}231.5 \\
    \bottomrule
    \end{tabular}
    }
\end{table}

To scrutinize performance over the forecast horizon, Table \ref{tab:motion_eval} presents a time-step-level analysis on the MOCAP-UMPM and 3DPW datasets. The results reveal not only the consistent superiority of our models over T2P and EMPMP at every interval but also a crucial architectural trade-off.
Our ``wide'' model establishes a new standard for local pose accuracy (APE), excelling at capturing fine-grained kinematics, particularly in the short term. Conversely, our ``deep'' model demonstrates its strength in long-range forecasting, achieving the best overall world-coordinate accuracy (JPE) at the final timesteps. This divergence highlights a key finding: architectural depth appears more critical for maintaining global trajectory coherence, while width is more effective for local pose detail. Most notably, the performance gap between our models and the baselines widens as the prediction horizon increases. This demonstrates our architecture's superior robustness against the error accumulation that typically plagues sequential prediction tasks. 

\subsection{Architectural Component Ablation}
\begin{table}[t]
    \centering
    \caption{Extended ablation study of architecture choices on MOCAP-UMPM. We explicitly compare RMSNorm vs. LayerNorm and the effect of Type Embeddings. Lower values are better ($\downarrow$), with the best results shown in \textbf{bold}.}
    \label{tab:ablation_extended}
    \small
    \begin{tabular}{ccccccc}
        \toprule
        \multicolumn{3}{c}{\textbf{Components}} & \multicolumn{2}{c}{\textbf{Ours (wide)}} & \multicolumn{2}{c}{\textbf{Ours (deep)}} \\
        \cmidrule(lr){1-3} \cmidrule(lr){4-5} \cmidrule(lr){6-7}
        \textbf{RMSNorm} & \textbf{LayerNorm} & \textbf{Type Emb.} & \textbf{APE}$\downarrow$ & \textbf{JPE}$\downarrow$ & \textbf{APE}$\downarrow$ & \textbf{JPE}$\downarrow$ \\
        \midrule
                   & \checkmark & \checkmark & 126.24 & 213.85 & 132.09 & 212.13 \\
        \checkmark &            &            & 127.37 & 213.06 & 132.80 & 214.85 \\
        \checkmark &            & \checkmark & \textbf{125.70} & \textbf{212.72} & \textbf{131.41} & \textbf{211.76} \\
        \bottomrule
    \end{tabular}
\end{table}

\begin{table}[t]
    \centering
    \caption{Ablation study on attention mechanisms measuring metrics on MOCAP-UMPM. Lower values are better $(\downarrow)$, with the best results shown in \textbf{bold}.}
    \label{tab:attn_ablation}
    \small
    \begin{tabular}{lcccc}
        \toprule
        & \multicolumn{2}{c}{\textbf{Wide}} & \multicolumn{2}{c}{\textbf{Deep}} \\
        \cmidrule(lr){2-3} \cmidrule(lr){4-5}
        \textbf{Attention Mechanism} & \textbf{APE}$\downarrow$ & \textbf{JPE}$\downarrow$ & \textbf{APE}$\downarrow$ & \textbf{JPE}$\downarrow$\\
        \midrule
        Self-Attn over [$\mathcal{C}; \mathcal{Q}$] (Ours) & \textbf{125.70} & \textbf{212.72} & \textbf{131.41} & \textbf{211.76} \\
        Self-Attn ($\mathcal{C}$) + Cross ($\mathcal{Q}$) & 134.61 & 229.05 & 140.32 & 227.89 \\
        \bottomrule
    \end{tabular}
    \vspace*{-4mm}
\end{table}

\begin{wrapfigure}{r}{0.5\textwidth}
    \centering
    \includegraphics[width=\linewidth]{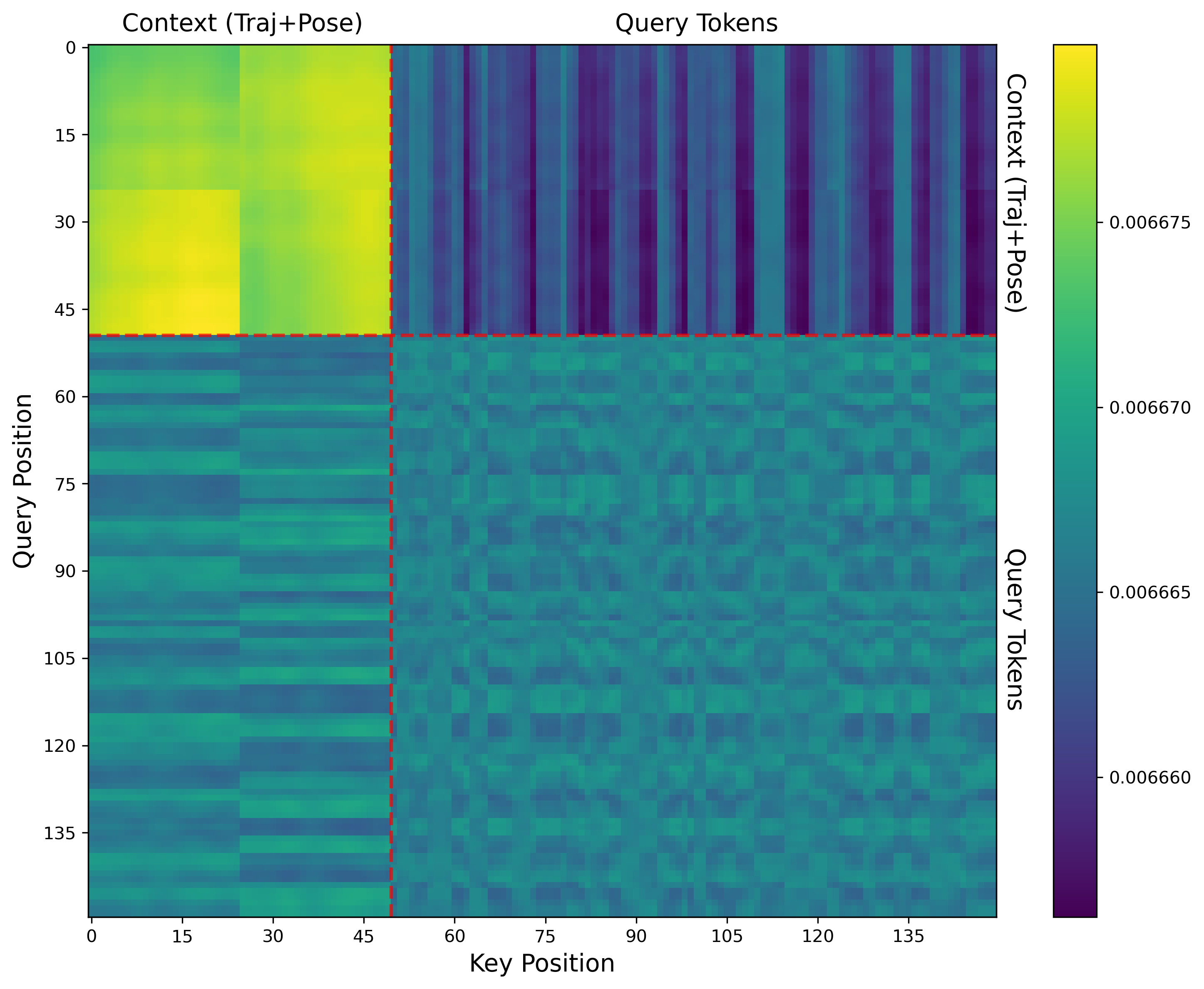}
    \caption{Attention patterns in the first transformer block at epoch 100. Brighter colors indicate stronger attention weights. Dashed lines (\textcolor{red}{- - -}) separate past context from query tokens.}
    \label{fig:attention_map}
    \vspace*{-4mm}
\end{wrapfigure}

We investigate the contribution of our smaller architectural choices. Table \ref{tab:ablation_extended} analyzes the impact of removing Type Embedding ($\epsilon$) and replacing RMSNorm with LayerNorm on the MOCAP-UMPM dataset. The results confirm their importance: removing the type embeddings degrades performance (\eg, APE increases from 125.70 to 126.24 for the wide model), confirming they are valuable for helping the model distinguish between pose and trajectory streams. RMSNorm also provides a consistent, though minor, benefit over LayerNorm while being more computationally efficient.

We further investigate the effectiveness of our unified self-attention mechanism, where past context and future queries are concatenated as ([$\mathcal{C}$; $\mathcal{Q}$]) and processed in a single block. This differs from standard encoder-decoders that use separate self-attention for context and cross-attention for queries. 

We evaluated a baseline replacing our unified attention with a standard encoder-decoder design, keeping all other parameters identical. As detailed in Table \ref{tab:attn_ablation}, our unified mechanism outperforms the cross-attention baseline, improving APE by 6.6\% and JPE by 7.1\% for the wide model. 

Visual analysis of attention patterns in the first transformer layer (Fig. \ref{fig:attention_map}) shows significant attention weights not only for queries that attend to past context (bottom-left quadrant) but also for queries that attend to other future queries (bottom-right quadrant). This confirms that the model learns complex, bidirectional relationships that a standard one-way query-to-context flow would fail to capture.

\subsection{Results on NBA SportVU}

To specifically assess the model's generalization capability in challenging multi-agent interactions, we conducted an evaluation on the NBA SportVU dataset, predicting 4 seconds of trajectory from 2 seconds of history. Unlike the standard pedestrian dynamics in ETH-UCY or social mingling in UMPM, NBA features high-intensity sports trajectories (basketball) characterized by rapid changes in velocity and complex interactions. This serves as a rigorous stress test for our architecture in scenarios where prior work typically relies on dedicated interaction modules.
\begin{table}[t]
    \centering
    \caption{Trajectory prediction performance on the NBA dataset. Lower values are better $(\downarrow)$, with the best results shown in \textbf{bold}.}
    \label{tab:nba_results}
    \small
    \setlength{\aboverulesep}{0pt}
    \setlength{\belowrulesep}{0pt}
    \begin{tabular}{l >{\columncolor{trajcolor}}c>{\columncolor{trajcolor}}c>{\columncolor{trajcolor}}c>{\columncolor{trajcolor}}c >{\columncolor{trajcolor}}c>{\columncolor{trajcolor}}c>{\columncolor{trajcolor}}c>{\columncolor{trajcolor}}c}
    \toprule
    & \multicolumn{4}{>{\columncolor{trajcolor}}c}{\textbf{ADE ($\downarrow$)}} & \multicolumn{4}{>{\columncolor{trajcolor}}c}{\textbf{FDE ($\downarrow$)}} \\
    \cmidrule(lr){2-5} \cmidrule(lr){6-9}
     & \textbf{1.0s} & \textbf{2.0s} & \textbf{3.0s} & \textbf{4.0s} & \textbf{1.0s} & \textbf{2.0s} & \textbf{3.0s} & \textbf{4.0s} \\
    \midrule
    Social-GAN    & 0.41 & 0.81 & 1.19 & 1.59 & 0.62 & 1.32 & 1.94 & 2.41 \\
    Social-STGCNN & 0.34 & 0.71 & 1.09 & 1.53 & 0.48 & 0.94 & 1.77 & 2.26 \\
    PECNet        & 0.40 & 0.83 & 1.27 & 1.69 & 0.71 & 1.61 & 2.44 & 2.95 \\
    STAR          & 0.43 & 0.75 & 1.03 & 1.13 & 0.66 & 1.24 & 1.51 & 2.01 \\
    Trajectron++  & 0.30 & 0.59 & 0.85 & 1.15 & 0.38 & 0.82 & 1.24 & 1.57 \\
    MemoNet       & 0.38 & 0.71 & 1.00 & 1.25 & 0.56 & 1.14 & 1.57 & 1.47 \\
    NPSN          & 0.35 & 0.68 & 1.01 & 1.31 & 0.58 & 1.23 & 1.76 & 1.79 \\
    GroupNet      & 0.26 & 0.49 & 0.73 & 0.96 & 0.34 & 0.70 & 1.02 & 1.30 \\
    MID           & 0.28 & 0.51 & 0.71 & 0.96 & 0.37 & 0.72 & 0.98 & 1.27 \\
    LED           & 0.18 & 0.37 & 0.58 & 0.82 & 0.27 & 0.56 & 0.85 & 1.15 \\
    NMRF          & \textbf{0.16 }& \textbf{0.34} & \textbf{0.53} & \textbf{0.75} & \textbf{0.24}& \textbf{0.50} & \textbf{0.75} & \textbf{0.97} \\
    \rowcolor{gray!20} Ours (wide) & 0.26 & 0.48 & 0.61 & 0.89 &  0.32 & 0.68 & 0.96 & 1.26\\
    \rowcolor{gray!20} Ours (deep) & 0.25 & 0.47 & 0.61 & 0.88 &  0.30 & 0.66 & 0.95 & 1.24\\
    \bottomrule
    \end{tabular}
    \vspace*{-4mm}
\end{table}

As shown in Table~\ref{tab:nba_results}, our model demonstrates strong and competitive performance across all prediction horizons, with the deep variant consistently outperforming the wide variant. Our approach surpasses several well-established baselines that explicitly model agent interactions, including GroupNet~\citep{groupnet} and MID~\citep{gu2022stochastic}, across all prediction horizons.
However, NMRF remains the top-performing method on this benchmark, establishing a clear performance ceiling that our interaction-free architecture does not fully close. 
These results further reinforce our earlier observation that incorporating a lightweight interaction module into our framework represents the most promising direction for future work, with the NBA setting serving as a compelling motivation for doing so. 

\section{Joint Training}
\label{sec:app:joint}
To test the full generalization capability of our architecture, we train a single, universal model jointly across all datasets and tasks (pose, trajectory, and combined prediction). This experiment aims to create a single set of weights that can perform any specialized task without retraining. Handling the significant diversity in data formats, skeletal structures, and sequence lengths requires a carefully designed methodology that we detail below.

\subsection{Methodology}
\paragraph{Data Unification and Canonical Skeleton.}
A primary challenge is the heterogeneity of the datasets. To create a consistent input format, all data is preprocessed into a normalized tensor of shape $T \times M \times 3$ (sequence length $\times$ joints $\times$ coordinates). We pad the data with a zero Z-dimension for 2D trajectory datasets (ETH-UCY, SDD) to create a consistent 3D representation.

\begin{table}[t]
\centering
\caption{Mapping from dataset-specific skeletons to our 22-joint canonical representation. AMASS serves as the canonical skeleton itself. Dashes (--) indicate that a direct mapping for that specific canonical joint is unavailable in the source dataset.}
\label{tab:skeleton_mapping}
\small
\begin{tabular}{@{}cllll@{}}
\toprule
\textbf{\#} & \textbf{AMASS} & \textbf{Human3.6M} & \textbf{MOCAP-UMPM} & \textbf{3DPW} \\ \midrule
1 & Pelvis & -- & Hips & Pelvis \\
2 & L\_Hip & LeftUpLeg & LHip & LHip \\
3 & R\_Hip & RightUpLeg & RHip & RHip \\
4 & Spine1 & Spine & Spine & -- \\
5 & L\_Knee & LeftLeg & LKnee & LKnee \\
6 & R\_Knee & RightLeg & RKnee & RKnee \\
7 & Spine2 & -- & -- & -- \\
8 & L\_Ankle & LeftFoot & LAnkle & -- \\
9 & R\_Ankle & RightFoot & RAnkle & -- \\
10 & Spine3 & -- & -- & -- \\
11 & L\_Foot & -- & -- & LFoot \\
12 & R\_Foot & -- & -- & RFoot \\
13 & Neck & Neck & Neck & -- \\
14 & L\_Collar & -- & -- & -- \\
15 & R\_Collar & -- & -- & -- \\
16 & Head & Head / Head-top & Head & -- \\
17 & L\_Shoulder & LeftArm & LShoulder & LShoulder \\
18 & R\_Shoulder & RightArm & RShoulder & RShoulder \\
19 & L\_Elbow & LeftForeArm & LElbow & LElbow \\
20 & R\_Elbow & RightForeArm & RElbow & RElbow \\
21 & L\_Wrist & LeftHand & LWrist & LWrist \\
22 & R\_Wrist & RightHand & -- & RWrist \\ \bottomrule
\end{tabular}
\end{table}

To address the varying skeleton definitions, we establish a 22-joint canonical skeleton, using the AMASS dataset as our standard. All other datasets are mapped to this representation, as shown in Table~\ref{tab:skeleton_mapping}. This mapping allows us to use a fixed set of learnable joint embeddings, ensuring that input data for a given semantic body part (\eg, the `Left Knee') is always processed by its corresponding embedding, regardless of the source dataset. For trajectory-only datasets, the single trajectory point is mapped to the `Pelvis' joint embedding.

\paragraph{Dataset-Balanced Batching.}
We employ a dataset-balanced batching strategy to prevent the model from overfitting to larger datasets (\eg, AMASS). Each training batch contains samples drawn from only a single dataset. We iterate through an equal number of batches from every dataset during each epoch, ensuring the model is exposed to a balanced distribution of tasks and data sources during training.

\paragraph{Task-Specific Processing.}
We use a task-type flag associated with each dataset to direct samples through the appropriate processing pipelines. For instance, a `trajectory' flag ensures that data only passes through the trajectory-related input and output heads of the model, while a `joint' flag activates both pose and trajectory heads. This allows the shared transformer core to learn a general motion representation while the specialized heads handle the task-specific details.

\paragraph{Unified Model with Dynamic Slicing.}
The model's internal parameters are defined by the maximum sequence length, $\max(T)$, and maximum number of joints, $\max(M)$, across all datasets. However, at runtime, a given sample's input and output tensors are dynamically sliced to match the specific $T$ and $M$ of its source dataset. This allows a single, fixed-size model to efficiently process variable-dimension inputs and outputs.

\subsection{Results}
\begin{table}[t]
\centering
\caption{Comparison of performance on individual vs.\ joint training. Lower values are better $(\downarrow)$.} 
\label{tab:joint_training_table}
\small
\setlength{\aboverulesep}{0pt}
\setlength{\belowrulesep}{0pt}
\renewcommand\theadalign{cc} 
\renewcommand\theadfont{\bfseries} 
\begin{adjustbox}{width=\linewidth}
\begin{tabular}{l>{\columncolor{posecolor}}c>{\columncolor{posecolor}}c
                >{\columncolor{trajcolor}}c>{\columncolor{trajcolor}}c
                >{\columncolor{bothcolor}}c>{\columncolor{bothcolor}}c}
\toprule
& \multicolumn{2}{>{\columncolor{posecolor}}c}{\textbf{Pose Prediction}} 
& \multicolumn{2}{>{\columncolor{trajcolor}}c}{\textbf{Trajectory Prediction}}
& \multicolumn{2}{>{\columncolor{bothcolor}}c}{\textbf{Pose + Trajectory Prediction}} \\
\cmidrule(lr){2-7} 
\thead{Dataset \\ In/Out length (s) \\ Metric} 
& \thead{Human3.6M \\ 0.5/2.0 \\ ADE$\downarrow$/FDE$\downarrow$} & \thead{AMASS \\ 0.5/2.0 \\ ADE$\downarrow$/FDE$\downarrow$}
& \thead{ETH-UCY (Avg) \\ 3.2/4.8 \\ ADE$\downarrow$/FDE$\downarrow$} & \thead{SDD \\ 3.2/4.8 \\ ADE$\downarrow$/FDE$\downarrow$}
& \thead{MOCAP-UMPM \\ 1.0/2.0 \\ APE$\downarrow$/JPE$\downarrow$} & \thead{\makebox[6.5em][c]{3DPW} \\ 0.8/1.6 \\ APE$\downarrow$/JPE$\downarrow$ } \\
\midrule
Ours (wide, ind.) & 0.42/0.59 & 0.31/0.45 & 0.18/0.32 & 6.70/7.63 & 125.70/212.72 &142.89/230.97\\
Ours (deep, ind.) &0.44/0.57 & 0.35/0.47& 0.19/0.32& 6.26/7.61 &131.41/211.76 &148.91/231.48 \\
Ours (wide, joint) & 0.49/0.63 &  0.51/0.66 & 0.23/0.37 &9.04/11.21& 135.19/220.13&150.40/234.81\\
Ours (deep, joint) & 0.55/0.70 & 0.62/0.78 & 0.25/0.39 & 10.66/12.14 & 138.20/223.49 & 151.46/235.05 \\
\bottomrule
\end{tabular}
\end{adjustbox}
\end{table}

The results of our joint training experiment, presented in Table \ref{tab:joint_training_table}, demonstrate both the promise and the challenges of creating a single, universal motion prediction model. As expected, there is a performance trade-off when compared to the specialized models trained on individual datasets. The jointly trained models exhibit a degradation in accuracy across all tasks and datasets. However, the degree of this degradation varies, providing valuable insights into the model's behavior.

The ``wide'' model consistently outperforms the ``deep'' model in the joint training setting. This is the inverse of our findings in some specialized tasks, and it suggests that the higher parameter count and wider embedding dimension of the ``wide'' model provide the necessary capacity to learn a shared representation across the seven diverse datasets. The ``deep'' model, with its constrained architecture, likely lacks the capacity to effectively generalize across such a heterogeneous data distribution, leading to a more significant performance drop. We also observe that the performance degradation is most pronounced on the AMASS dataset. This is likely a direct consequence of our dataset-balanced batching strategy. While this strategy prevents the model from overfitting to the largest datasets, it also means that the model is significantly under-exposed to the vast and diverse AMASS dataset, which is over 140 times larger than the smallest dataset (SDD). The model simply does not see enough of the AMASS data distribution to learn it as effectively as the specialized model.

Despite the performance trade-off, these results represent a successful proof of concept. The ability of a single, simple architecture to perform pose prediction, trajectory forecasting, and combined holistic prediction without any architectural changes is a powerful demonstration of its inherent generality. The fact that the model produces reasonable, albeit less accurate, predictions across all tasks indicates that it has learned a meaningful and transferable internal representation of human motion. This experiment validates the potential for developing true ``foundation models for motion.'' While our current approach shows a performance gap, it highlights a clear and promising research direction. Future work could focus on more sophisticated data-balancing techniques, curriculum learning strategies, or scaling the model's capacity to bridge this gap. The ability to train a single model that understands the principles of human motion across myriad contexts remains a valuable and achievable goal for the field.

\section{Evaluating Prediction Diversity}
\label{sec:app:diversity}

A key component of our model is the multi-modal prediction head, which generates $K$ distinct hypotheses to account for the uncertain nature of human motion. To validate its effectiveness, we analyze two potential concerns: mode collapse and the true diversity of the generated futures.

\begin{figure}[t]
    \centering
    \begin{subfigure}[b]{0.48\textwidth}
        \centering
        \includegraphics[width=\textwidth]{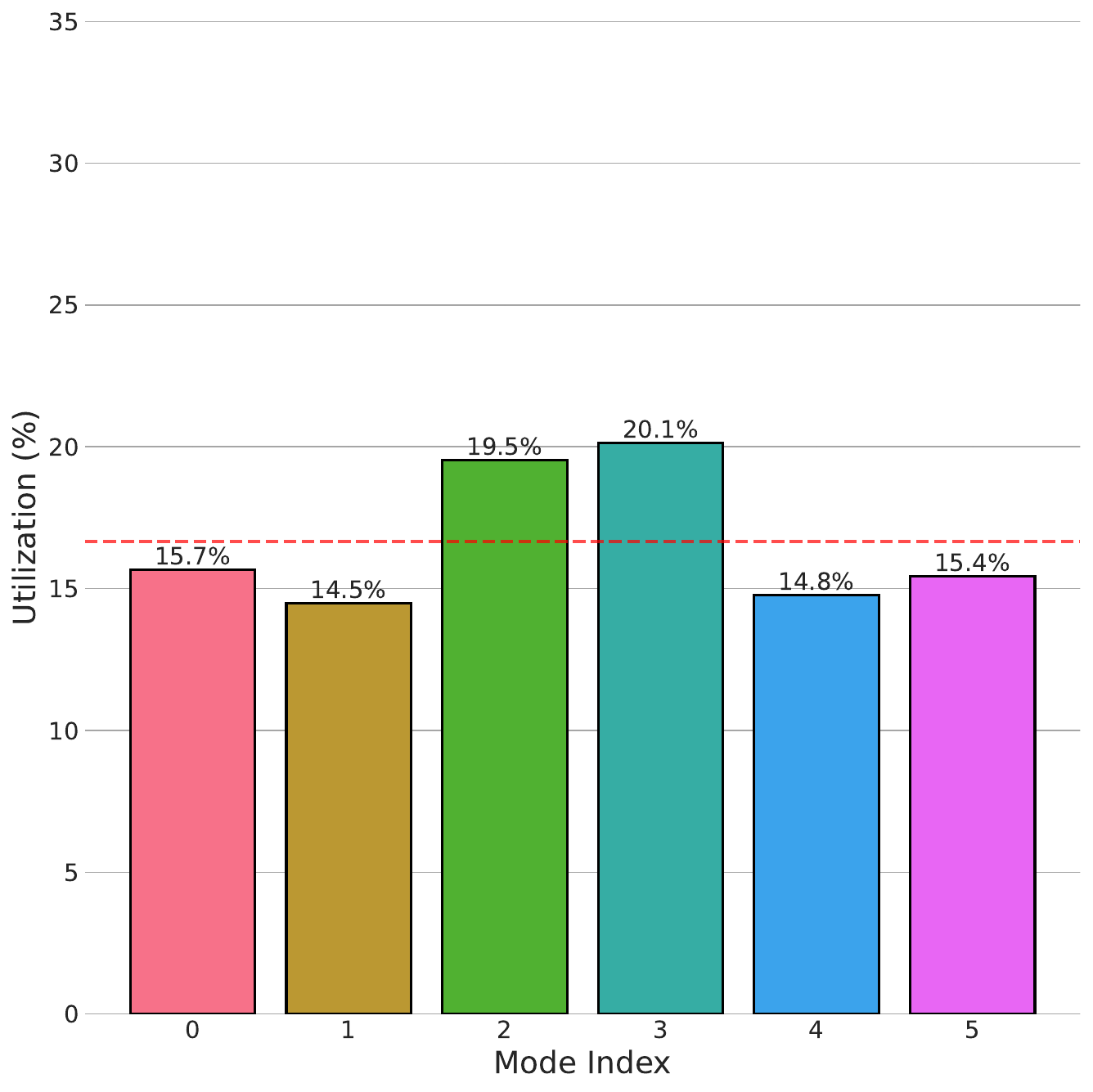}
        \caption{Training set}
        \label{fig:mode_dist_training}
    \end{subfigure}
    \hfill
    \begin{subfigure}[b]{0.48\textwidth}
        \centering
        \includegraphics[width=\textwidth]{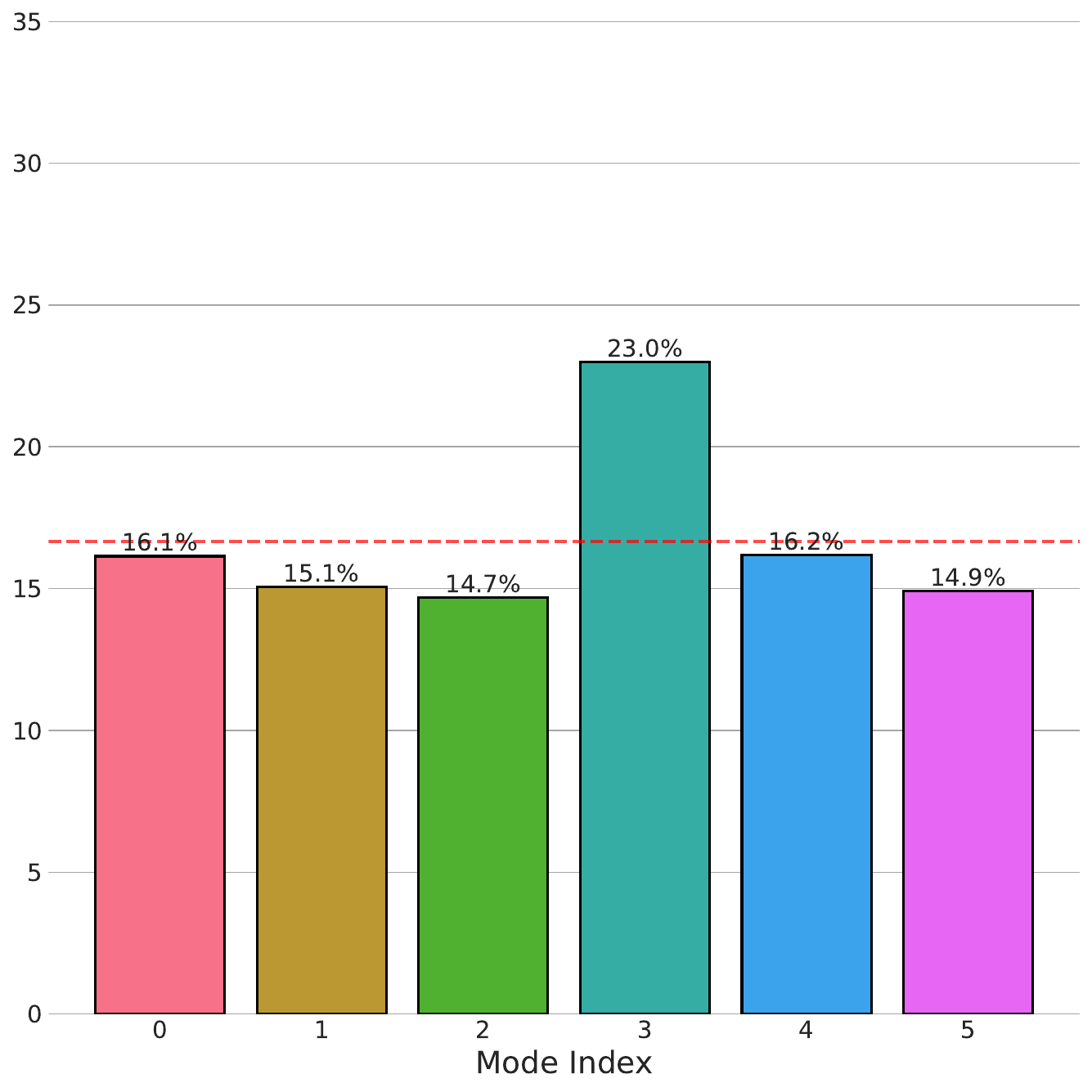}
        \caption{Validation set}
        \label{fig:mode_dist_validation}
    \end{subfigure}
    \caption{Distribution of winning mode indices (best-of-6) on pose + trajectory prediction task across the training and validation sets of MOCAP-UMPM. The dashed line (\textcolor{red}{- - -}) indicates equal distribution. Both distributions demonstrate balanced mode utilization without mode collapse.}
    \label{fig:mode_distribution}
    \vspace*{-4mm}
\end{figure}

\subsection{Mode Utilization Analysis}
To quantitatively verify that our model uses its full predictive capacity, we logged the index of the best (lowest error) hypothesis for every sample in the MOCAP-UMPM training and validation sets ($K=6$). Figure \ref{fig:mode_distribution} plots the distribution of these winning indices. The results show that the model does not suffer from mode collapse. All six modes are actively utilized in both training and validation, with utilization rates clustering around the ideal uniform distribution (16.7\%, shown as a dashed line). This confirms that the ``winner-takes-all'' loss, when applied to our architecture, successfully encourages the different proposals to specialize and cover distinct, plausible outcomes.

\begin{table}[t]
\centering
\caption{Comparison of pose prediction diversity. Lower values are better $(\downarrow)$, with the best results shown in \textbf{bold}. An asterisk (*) denotes models we recomputed for this setup.}
\label{tab:pose_diversity}

\renewcommand\theadalign{cc}
\renewcommand\theadfont{\bfseries}

\begin{adjustbox}{width=\linewidth}
\begin{tabular}{ll
                >{\columncolor{posecolor}}c
                >{\columncolor{posecolor}}c
                >{\columncolor{posecolor}}c
                >{\columncolor{posecolor}}c}
\toprule
& & \multicolumn{2}{>{\columncolor{posecolor}}c}{\textbf{Human3.6M}} 
& \multicolumn{2}{>{\columncolor{posecolor}}c}{\textbf{AMASS}} \\
\cmidrule(lr){3-4} \cmidrule(lr){5-6} 
\thead{Model} & \thead{Type}&
\thead{MMADE$\downarrow$} & 
\thead{MMFDE$\downarrow$} &
\thead{MMADE$\downarrow$} & 
\thead{MMFDE$\downarrow$} \\ 
\midrule

DivSamp    & Stochastic (Gumbel-Softmax) & 0.542          & 0.671         & 0.623 & 0.728 \\
BeLFusion  & Stochastic (Latent Diffusion) & 0.491   & 0.586    & 0.488 & 0.564 \\
CoMusion   & Stochastic (Motion Diffusion) & 0.531  & 0.623        & 0.526 & 0.602 \\
Motionmap  & Stochastic (Multi-Stage Encoder-Decoder) & \textbf{0.466}   & \textbf{0.532}   &  \textbf{0.450} & \textbf{0.514} \\
SkeletonDiff* &Stochastic (Gaussian Diffusion)& 0.568 & 0.694 & 0.641 & 0.740 \\
SLD*          &Stochastic (State-Space Diffusion)& 0.497 & 0.576 & 0.482 & 0.551 \\
\cellcolor{highlightgray}Ours (wide)  &\cellcolor{highlightgray}Deterministic (K-Proposal)& \cellcolor{highlightgray}0.526   & \cellcolor{highlightgray}0.587 & \cellcolor{highlightgray}0.519 & \cellcolor{highlightgray}0.560 \\
\cellcolor{highlightgray}Ours (deep)  &\cellcolor{highlightgray}Deterministic (K-Proposal)& \cellcolor{highlightgray}0.535    & \cellcolor{highlightgray}0.597    & \cellcolor{highlightgray}0.521        & \cellcolor{highlightgray}0.571\\
\bottomrule
\end{tabular}
\end{adjustbox}
\end{table}

\subsection{Quantitative Diversity Metrics}
To further assess the quality and diversity of our generated motion distribution, we report standard multi-modal metrics, Minimum-over-K Average Displacement Error (MMADE) and Minimum-over-K Final Displacement Error (MMFDE), on the pose-only benchmarks. Table \ref{tab:pose_diversity} compares our model's performance against prominent stochastic and generative baselines. Our deterministic K-proposal approach achieves MMADE/MMFDE scores competitive with those of these generative methods, demonstrating that our $K$ proposals capture a meaningful and diverse set of high-quality future motions. We additionally measure Average Pairwise Distance (APD; higher is more diverse) on Human3.6M. Our deep model obtains an APD of 7.751, compared with 7.632 for CoMusion, confirming substantial variation across its proposals.

\subsection{Selecting One Hypothesis}
The model produces $K$ unordered hypotheses. For applications requiring one prediction, training with $K=1$ yields an APE/JPE of 145.84/280.8 on MOCAP-UMPM (Table~\ref{tab:k_comparison}), outperforming the single-output T2P and EMPMP baselines. We also tested a linear scoring head over the $K$ branches, supervised by cross-entropy against the winner-takes-all index. Its highest-scoring hypothesis achieves a single-output JPE of 279.6, compared with 280.8 for the $K=1$ model. Joint training with the scoring objective yields a best-of-$K$ JPE of 224.1, compared with 212.7 for winner-takes-all training. These results characterize the trade-off between selecting a broadly reliable hypothesis and specializing proposals toward distinct futures.

\section{Additional Qualitative Results}
\label{sec:app:qual}
\subsection{Trajectory-Only Visualization}
\begin{figure}[t]
    \centering     
    \begin{subfigure}[b]{0.49\textwidth}
        \centering
        \includegraphics[width=\textwidth]{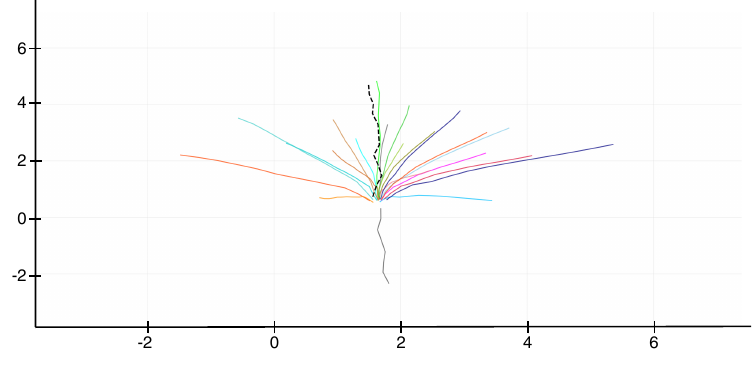}
    \end{subfigure}
    \hfill
    \begin{subfigure}[b]{0.49\textwidth}
        \centering
        \includegraphics[width=\textwidth]{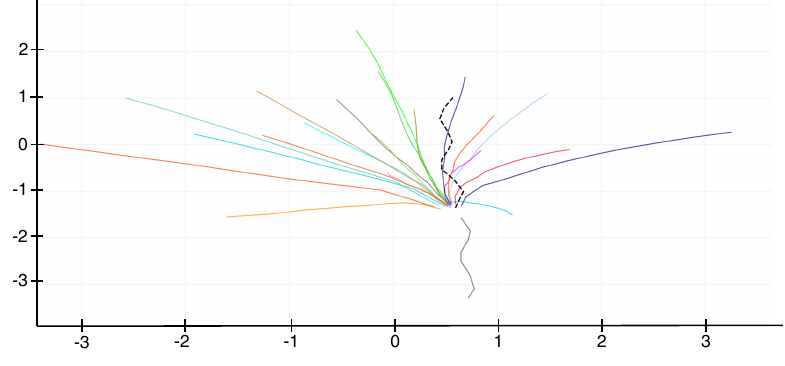}
    \end{subfigure}
    \medskip 
    \begin{subfigure}[b]{0.49\textwidth}
        \centering
        \includegraphics[width=\textwidth]{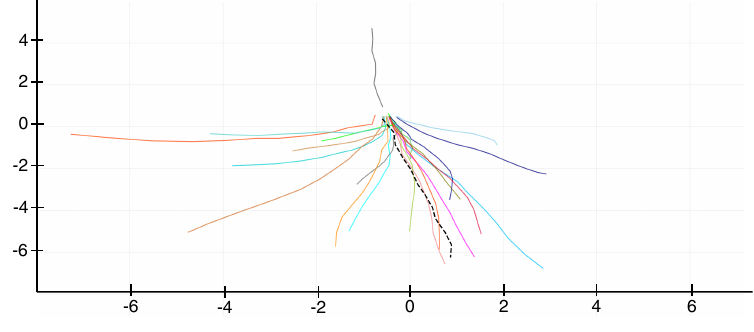}
    \end{subfigure}
    \hfill
    \begin{subfigure}[b]{0.49\textwidth}
        \centering
        \includegraphics[width=\textwidth]{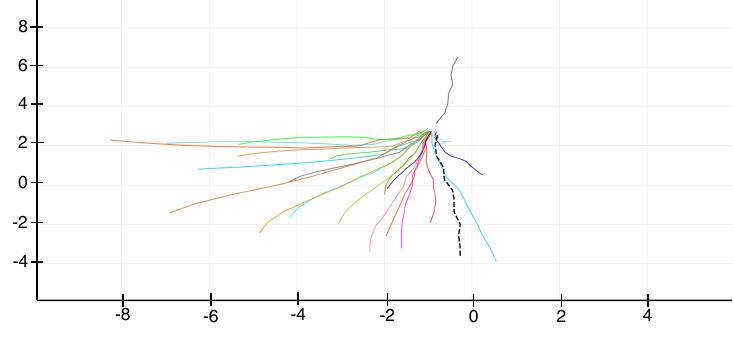}
    \end{subfigure}
    \caption{Visualization of trajectory predictions ($K = 20$) of our (wide) model on ETH-UCY data. X-Y coordinate values in the plots are in meters. All model predictions are in color, ground truth future trajectories are black dashes, and input trajectories are in gray.}
    \label{fig:traj_viz}
\end{figure}
To provide qualitative results for the trajectory-only task, Figure \ref{fig:traj_viz} visualizes our model's $K=20$ proposals on challenging, crowded scenes from the ETH-UCY dataset. The visualizations show our model's ability to capture a wide, multi-modal distribution of plausible future paths, correctly identifying diverse outcomes (\eg, turning left, turning right, or stopping) in high-uncertainty scenarios.

\subsection{Failure Case Analysis}
\begin{figure}[t]
    \centering
    \includegraphics[width=\linewidth]{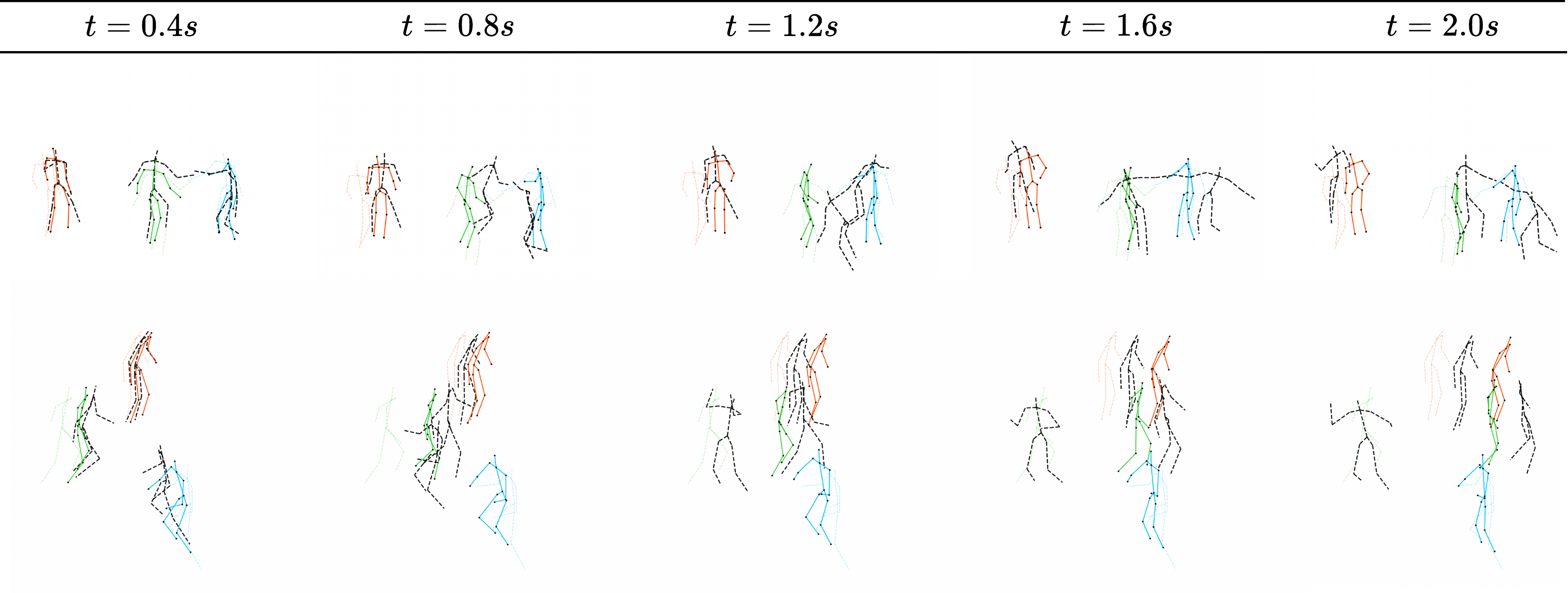}
    \caption{Visualization of predictions of our (wide) model on MOCAP-UMPM data. All model predictions are in color. Ground truth future poses are black dashes, and the last-known input positions are colored dashes.}
    \label{fig:failure_viz}
\end{figure}

While our model is robust, it is not without limitations, particularly in complex multi-person scenes. Our model treats all individuals independently, which can lead to unrealistic predictions when agents' motions are strongly coupled or highly unusual. Figure \ref{fig:failure_viz} presents qualitative failure cases from MOCAP-UMPM. The first example showcases an intricate interaction in which the blue and green agents turn in circles while holding hands. Our model fails to capture this complex, coupled motion. This also highlights that integrating explicit multi-agent interaction modules is a critical and promising direction for future work. The second example shows the blue and green agents undergoing an unexpected, rapid acceleration. The model's predictions struggle to keep pace with this abrupt change in dynamics, likely defaulting to a smoother, more mean-reverting trajectory and thus accumulating significant error.

\section{Website}
\label{sec:app:web}
We provide a website with additional visualizations demonstrating our method's performance, which can be accessed using the provided HTML file. We observe that the generated motions exhibit high physical plausibility, with no unrealistic artifacts such as foot sliding. Body poses are consistently realistic, respecting natural body constraints and capturing fine-grained details without grouping different joints into unnatural, blocky movements. Furthermore, our model adeptly handles both independent and coupled motion dynamics; it accurately predicts localized movements (\eg, arm gestures without a change in trajectory) and complex actions where limb articulation and global trajectory are deeply intertwined. 

\section{Licenses}
\label{sec:app:license}
All datasets used in this work are publicly available and used in accordance
with their respective licenses. AMASS is released under a CC BY 4.0 license.
Human3.6M is available under a custom research license restricted to
non-commercial use. ETH-UCY, SDD, MOCAP-UMPM, and 3DPW are available for
non-commercial research use only. All datasets are used strictly in accordance
with these terms.

\end{document}